\documentclass{article}

\usepackage[final]{corl_2026} 

\usepackage{amsmath}
\usepackage{amssymb}
\usepackage{graphicx}
\usepackage{booktabs} 
\usepackage{colortbl} 
\usepackage{xcolor}
\usepackage{multirow}
\usepackage{caption}
\usepackage{wrapfig}
\usepackage{placeins}
\usepackage{algorithm}
\usepackage{algpseudocode}
\usepackage{amsmath}
\usepackage{hyperref}

\algrenewcommand\algorithmiccomment[1]{\hfill $\triangleright$ #1}

\title{ProxiDex: Learning Dynamics-Guided Proximity Policy for Dexterous Manipulation}

\author{
	Yushan Bai$^{1,2,*}$ \quad
	Boyu Zheng$^{1,2,*}$ \quad
	Zhiyang Mao$^{4}$ \quad
	Hongzheng Sun$^{1,2}$ \\
	\textbf{Yuchuang Tong}$^{1,2,\dagger}$ \quad
	\textbf{En Li}$^{1,2,3}$ \quad
	\textbf{Zhengtao Zhang}$^{1,2,3,\dagger}$ \\[3pt]
	{\small $^{1}$CAS Engineering Laboratory for Intelligent Industrial Vision, Institute of Automation, Chinese Academy of Sciences} \\
	{\small $^{2}$School of Artificial Intelligence, University of Chinese Academy of Sciences} \\
	{\small $^{3}$Beijing Zhongke Huiling Robot Technology Co., LTD.} \\
	{\small $^{4}$School of Intelligent Science and Technology, Xinjiang University} \\[2pt]
	{\small $^{*}$Equal contribution. \quad
		$^{\dagger}$Corresponding authors.}
}

\begin{document}
\maketitle


\begin{figure}[!htbp]
	\centering
	\captionsetup{skip=4pt}
	\includegraphics[width=\linewidth]{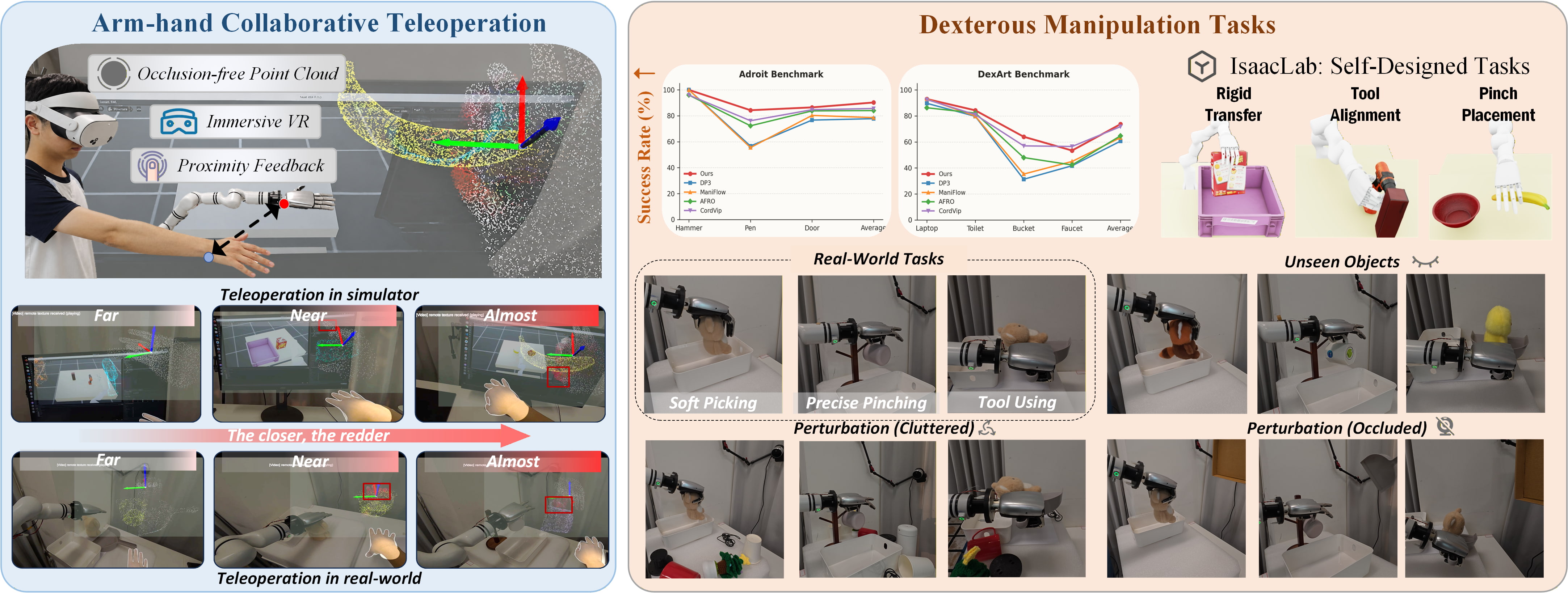}
	\caption{We propose ProxiDex, a framework that integrates teleoperation with proximity-policy learning. The left panel shows immersive teleoperation with proximity feedback, while the right panel demonstrates robust performance in both simulation and real-world dexterous manipulation.}
	\label{fig1}
	\vspace{-10pt}
\end{figure}

\begin{abstract}
    Multi-finger dexterous manipulation relies on stable hand-object interactions, yet these interactions are partially observable in practice. Visual observations are often occluded by the hand, tactile sensors introduce hardware-specific modalities and calibration burdens, and existing policies rarely model how these cues evolve under actions, making them brittle under contact uncertainty. To address these, we present ProxiDex, a dynamics-guided proximity policy framework that treats hand-object proximity as an interaction state for dexterous manipulation. ProxiDex reconstructs interaction point clouds and converts geometric distances into proximity cues, forming a hardware-agnostic contact representation that provides immersive feedback during VR teleoperation. Built on this representation, ProxiDex learns action-conditioned proximity dynamics with a coupled forward–inverse design: future observation latents are predicted from actions, while proximity variations are decoded from latent changes. Leveraging these dynamics, ProxiDex adaptively reweights proximity tokens across manipulation phases and uses dynamics-consistency supervision to guide policy inference, stabilizing action generation under unreliable visual feedback. Simulation and real-world experiments demonstrate improved success rates and robustness over representative baselines across standard, unseen objects, and perturbation scenarios. Additional visualizations are available at \url{https://proxidex.github.io/}.
\end{abstract}

\keywords{Dexterous Manipulation, Interaction Dynamics, Imitation Learning}


\section{Introduction}
	
   Dexterous manipulation with multi-fingered hands is a fundamental capability for human-like robotic systems \cite{fu2025cordvip}. Compared with parallel jaw grippers, dexterous hands require coordinated control of arm-level motion and fingertip-level interactions to accomplish complex tasks such as grasping, insertion, and tool use \cite{bai2026far,qin2023dexpoint}. A key requirement for learning such skills is the ability to represent local hand-object geometric interaction, which governs how contact is formed, transferred, and maintained during manipulation \cite{xu2026contact}. However, acquiring such interaction cues from demonstrations remains challenging. Vision-based teleoperation systems \cite{handa2020dexpilot,iyer2025open} suffer from frequent occlusions that obscure critical contact regions, while tactile sensing provides limited scalability due to heterogeneous hardware designs and costly calibration across platforms \cite{huang20253d,ren2023mc}. As a result, reliable local interaction information is often missing or inconsistent in collected data, limiting the ability of visuomotor policies to infer the underlying and evolving physical interaction structure.
   
   This limitation exposes a broader gap in existing imitation learning approaches for dexterous manipulation. Most methods \cite{zhao2023learning, chi2025diffusion, ze2024dp3} directly learn end-to-end mappings from visual observations to actions, relying on implicit feature representations without explicitly modeling hand-object geometric proximity or its temporal evolution. Even when contact-related signals are incorporated, they are typically treated as auxiliary inputs without modeling how actions induce changes in interaction states or how interaction structure constrains future control \cite{li2025adaptive,11128094}. This leads to brittle behavior under occlusion, tracking noise, or contact uncertainty, where policies often generate physically inconsistent actions and fail to generalize beyond clean demonstrations.
   
   To address these challenges, we propose \textbf{ProxiDex}, a dynamics-guided proximity policy learning framework that builds on hand-object geometric proximity as a coherent interaction representation and models its action-conditioned evolution for robust policy learning. As shown in Fig.~\ref{fig1}, within a VR teleoperation setting \cite{weng2026hts}, we reconstruct interaction point clouds and visualize them as immersive proximity feedback for operators. We further convert point-cloud distances into hardware-agnostic proximity feedback, enabling scalable inference of local interaction cues without tactile instrumentation. Based on this representation, ProxiDex learns forward-inverse latent dynamics that couple action-driven observation transitions with proximity variations, capturing the evolution of local hand-object interactions during manipulation. Leveraging these interaction dynamics, ProxiDex adaptively reweights proximity tokens across manipulation phases, balancing global observation guidance during motion with local proximity constraints during interaction. Dynamics consistency further guides online inference, stabilizing action generation under occlusion and tracking failures.
   
   The main contributions of ProxiDex are summarized as follows: 1) We propose a data-construction pipeline and a VR-based teleoperation with proximity feedback, which provides scalable local interaction cues without tactile instrumentation. 2) We develop a forward-inverse latent dynamics model to capture action-conditioned evolution between observation transitions and proximity variations.  3) We introduce a trajectory-adaptive proximity policy with dynamics-consistency guided inference, improving robustness under occlusion and tracking failures.


\section{Related Work}
	
\textbf{Demonstration data acquisition for dexterous manipulation.}
High-quality demonstration data are fundamental to learning dexterous manipulation. Existing data-collection systems often trade off sensing accuracy against deployment cost \cite{zhao2023learning,fang2025airexo}. Methods based on data gloves or tactile sensing devices \cite{zhang2025doglove, wen2025dexterous, li2025haptic} can provide fine-grained contact feedback, but they typically rely on expensive hardware and require calibration, maintenance, and cross-platform alignment. Vision-based teleoperation methods \cite{iyer2025open,qin2023anyteleop,cheng2025open,ding2025bunny} reduce deployment barriers, yet they struggle to capture local proximity relations reliably under complex hand-object occlusions. UMI-style devices \cite{xu2025dexumi, tao2025dexwild} and video-parsing methods \cite{chen2026dexterous, mu2026deximit} offer promising scalability, but they are often constrained by specific end-effector designs or the quality of 3D reconstruction. In contrast, we adopt a VR-based arm-hand co-teleoperation pipeline and explicitly recover interaction structure through geometric proximity, enabling scalable and interaction-aware data collection.

\textbf{Imitation learning for dexterous manipulation.}
Imitation learning has advanced dexterous manipulation by modeling action distributions with Transformers or diffusion models \cite{zhao2023learning, chi2025diffusion}. Yet commonly used 2D observations struggle to capture the fine spatial relationships required for multi-finger contact. Point-cloud-based policies improve geometric awareness \cite{ze2024dp3, yan2025maniflow, fang2025flow}, but these methods still fail to maintain a consistent representation of interaction states under occlusion. Reconstruction and pose-based methods recover more complete geometry \cite{fu2025cordvip}, but mainly emphasize static geometric alignment rather than the temporal evolution of contact states. These limitations indicate that existing methods lack an explicit modeling of interaction as a structured and evolving state, motivating our explicit proximity-based interaction representation.

\textbf{Dynamics-aware representation learning.}
Using dynamics models to enhance the physical awareness of policies has become an important direction in dexterous manipulation. Latent dynamics pretraining methods \cite{liang2025bootstrap, lyu2026lda} improve the dynamics sensitivity of observation encoders by predicting future states, but their latent representations often lack explicit physical meaning. Multimodal world models \cite{zheng2026omnivta, higuera2026visuo, yuan2026vtam} assist action generation through future latent prediction, yet their complex model structures and inference pipelines can limit deployment efficiency. Geometry-evolution prediction method \cite{zheng2026emerging} can capture changes in point-cloud positions and velocities, but they remain limited in representing fine-grained contact logic in dexterous manipulation. Particle-based world models \cite{hong2025learning, he2025scaling} can reconstruct hand-object interaction processes, but they rarely model interaction explicitly as a bidirectional evolution of geometric proximity under actions. Motivated by these observations, we model interaction as a bidirectional evolution of geometric proximity under action influence, enabling explicit reasoning over contact dynamics.
	

\section{Problem Formulation}
Given an expert demonstration dataset $\mathcal{D}=\{\tau_i\}$, each trajectory $\tau=\{(o_t,a_t^E)\}_{t=1}^{T}$ consists of observations $o_t$ and expert actions $a_t^E$. Imitation learning aims to train a parameterized policy $\pi_\theta$ to match expert behavior. For 3D point-cloud observations, an encoder $f_\phi$ is typically used to map $o_t$ into a latent representation $z_t=f_\phi(o_t)$, from which the policy generates an action distribution $a_t \sim \pi_\theta(\cdot|z_t)$.

To improve the policy's awareness of future physical evolution, prior methods often introduce a latent dynamics model $h_\psi$ as auxiliary supervision. Given the current latent representation $z_t$ and a future $k$-step sequence of expert actions $a^E_{t:t+k-1}$, the model predicts the future latent state and is optimized through a feature reconstruction loss:
\begin{equation}
	\hat{z}_{t+k} = h_\psi(z_t, a^E_{t:t+k-1}),\quad \mathcal{L}_{dyn} = \mathbb{E}_{\tau \sim \mathcal{D}} [ \| \hat{z}_{t+k} - f_\phi(o_{t+k}) \|^2_2 ]
\end{equation}
However, directly modeling the observation latent space is insufficient for contact-rich dexterous manipulation. First, frequent hand-object occlusions cause the raw observation $o_t$ to miss fingertip contact regions, making interaction cues unobservable. Second, when only $\mathcal{L}_{dyn}$ is used to enforce future feature consistency, the model tends to prioritize salient motion patterns such as object-pose changes while overlooking the fine-grained evolution of hand-object distance and contact state. Our central motivation is therefore to construct a more physically meaningful interaction representation. We first reconstruct an occlusion-free hand-object interaction point cloud $o_t^p$ to mitigate geometric information loss. We then introduce latent interaction dynamics, consisting of a forward dynamics model and an inverse differential module, to help the policy anticipate proximity and contact-state evolution and generate reliable actions under perturbation.

\section{Method}

\begin{figure}[!t]
	\centering
	\captionsetup{skip=4pt}
	\includegraphics[width=\linewidth]{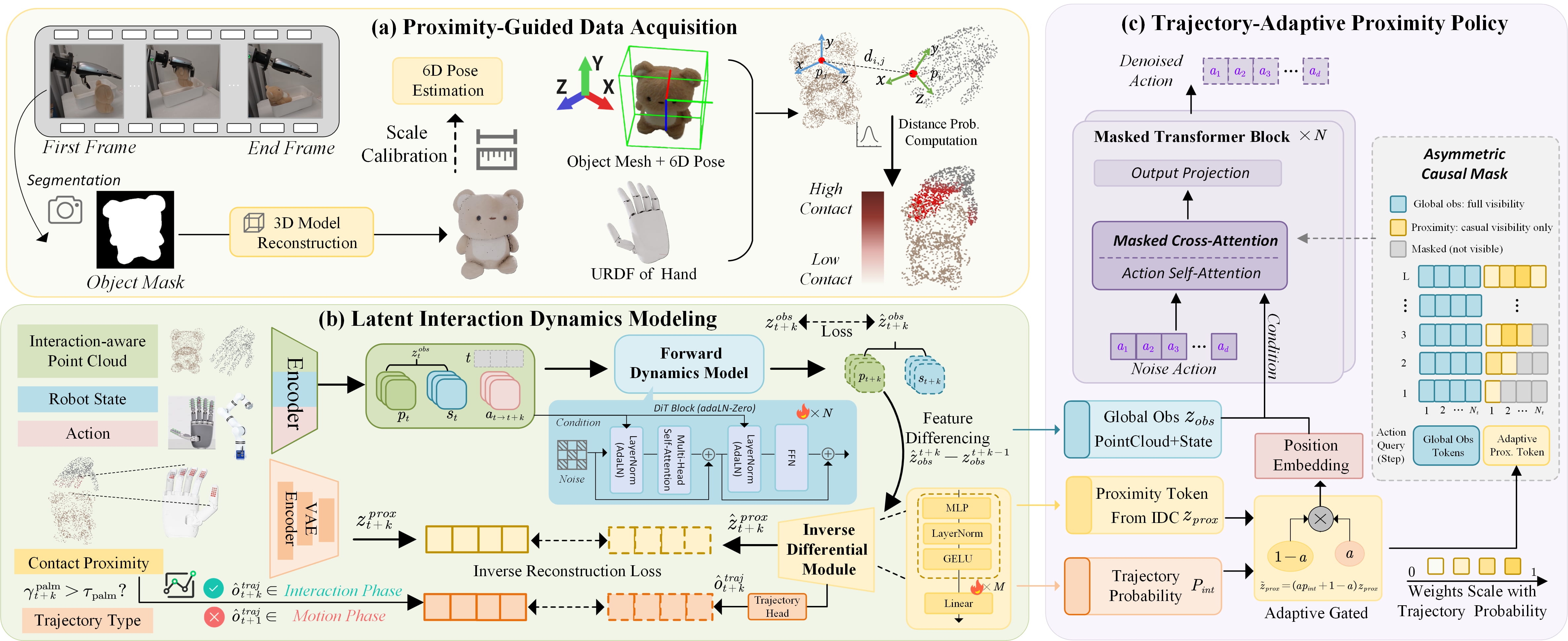}
	\caption{Overview of ProxiDex. It combines proximity-aware data acquisition, latent dynamics modeling, and trajectory-adaptive policy learning for robust dexterous manipulation. }
	\label{fig2}
	\vspace{-10pt}
\end{figure}

We propose \textbf{ProxiDex} to address visual occlusion and missing local interaction cues in dexterous manipulation, as illustrated in Fig.~\ref{fig2}. The method proceeds in three stages: (1) interaction-aware data acquisition, which reconstructs hand--object interaction point clouds from manipulation images and robot states (Sec.~\ref{sec:interaction_data_acquisition}); (2) latent interaction dynamics modeling, which extracts region-level proximity cues and learns action-conditioned proximity evolution (Sec.~\ref{sec:latent_interaction_dynamics}); and (3) trajectory-adaptive policy inference, which incorporates these cues into a diffusion policy to reweight proximity tokens across manipulation phases, while dynamics-consistency guidance supports closed-loop inference for robust action generation (Secs.~\ref{sec:trajectory_adaptive_policy} and~\ref{sec:dynamics_guided_inference}; Fig.~\ref{fig3}).

\subsection{Interaction-Aware Data Acquisition}
\label{sec:interaction_data_acquisition}

To collect high quality demonstrations that reflect hand-object contact relations, we build an interaction-aware data collection pipeline that unifies visual observations, robot proprioception, and proximity information in the robot base frame, as illustrated 
in Appendix Fig.~\ref{fig:vr_teleoperation}. Demonstrations are collected using a VR arm-hand co-teleoperation system based on Hand Tracking Streamer \cite{weng2026hts}, which synchronously records robot states, target-object poses, and expert actions. Details of teleoperation control, action retargeting, and proximity feedback in VR are provided in Appendix~\ref{app:vr_teleoperation}.

To represent hand-object interaction details, we first use Grounded-SAM \cite{ren2024grounded} and SAM3D \cite{chen2025sam} to reconstruct the target object from a single initial image, followed by metric scale calibration to obtain the object model $\mathcal{M}_o$, and then track its 6D pose during demonstrations using FoundationPose++ \cite{Wenhao_Yan_and_Jie_Chu_FoundationPose_2025}. The full procedure is described in Appendix~\ref{app:object_reconstruction}. Inspired by interaction-centric grasp representations \cite{wei2025mathcal}, we sample surface points from both the reconstructed object mesh and the hand-link at time $t$. As shown in Fig.~\ref{fig2}(a), the sampled points are then transformed into the robot base frame to obtain $P_t^o$ and $P_t^h$. The interaction point cloud observation at time $t$ is then defined as $o_t^p=P_t^o\cup P_t^h$. Compared with raw scene point clouds, the interaction point cloud preserves hand-object geometric relations more reliably and reduces the influence of background noise.

From $o_t^p$, we derive hand-object proximity observations by computing local distances between sampled hand surface points and object surface points. The resulting point-wise proximity strengths are then aggregated within each physical hand region $S_k$, yielding the region-level proximity vector $o_t^{\mathrm{prox}}=[\gamma_t^1,\dots,\gamma_t^K]$. The detailed computation is provided in Appendix~\ref{app:geometric_proximity}. In addition, we use the palm-region response $\gamma_t^{\mathrm{palm}}$ to generate a trajectory-phase label $o_t^{\mathrm{traj}}=\mathbb{I}(\gamma_t^{\mathrm{palm}}>\tau_{\text{palm}})$, where $o_t^{\mathrm{traj}}=0$ denotes the approach phase and $o_t^{\mathrm{traj}}=1$ denotes the contact-interaction phase. This label provides a coarse motion-phase prior for subsequent models. Each demonstration frame is finally represented as $(o_t^p,o_t^{\mathrm{prox}},o_t^{\mathrm{traj}},s_t,a_t)$, where $s_t$ is the robot proprioceptive state and $a_t$ is the expert action. This representation jointly contains the interaction point cloud, region-level proximity, and trajectory-phase label, providing a physically interpretable interaction observation for latent dynamics modeling and policy learning.

	\subsection{Latent Interaction Dynamics Modeling}
\label{sec:latent_interaction_dynamics}
After obtaining the interaction-aware demonstration representation, we model latent interaction dynamics with self-supervised learning, encouraging the observation encoder to capture action-driven changes in hand-object proximity.  As illustrated in Fig.~\ref{fig2}(b), we first pretrain a VAE on the proximity observation $o_t^{\mathrm{prox}}$, compressing region-level proximity responses into a proximity latent variable $z_t^{\mathrm{prox}}$ and reconstructing $\hat{z}_t^{\mathrm{prox}}$ to obtain a stable contact representation. This proximity latent variable then serves as a physical anchor for aligning visual and contact cues.

Next, the observation encoder $E_\theta$ encodes the interaction point cloud $o_t^p$ and robot state $s_t$ into an observation latent variable $z_t^{obs}$. A forward dynamics model $\mathrm{FDM}(\cdot)$ based on the DiT architecture \cite{peebles2023scalable} takes $z_t^{obs}$ and the action sequence $a_{t:t+h-1}$ as input and predicts future observation latents. Meanwhile, the inverse differential module $\mathrm{IDM}(\cdot)$ infers the proximity latent at the corresponding time step from predicted  latent variable differences and aligns it with the VAE-derived $z_t^{\mathrm{prox}}$. Details are provided in Appendix~\ref{app:latent_dynamics_architecture}:

\begin{equation}
	\small
	\mathcal{L}_{\mathrm{for}}
	=
	\frac{1}{H}
	\sum_{h=1}^{H}
	\|\mathrm{FDM}(z_t^{obs}, a_{t:t+h-1})-z_{t+h}^{obs}\|_2^2,
	\quad
	\mathcal{L}_{\mathrm{inv}}
	=
	\frac{1}{H}
	\sum_{h=1}^{H}
	\|\mathrm{IDM}(\hat{z}_{t+h}^{obs}-\hat{z}_{t+h-1}^{obs})-z_{t+h}^{\mathrm{prox}}\|_2^2
\end{equation}
 $\mathrm{FDM}(\cdot)$ constraint encourages $z_t^{\mathrm{obs}}$ to encode visual interaction and proprioceptive cues relevant to future state evolution, while  $\mathrm{IDM}(\cdot)$ alignment makes latent transitions sensitive to proximity variations. To distinguish trajectory phases, we introduce a trajectory head that predicts the interaction-phase probability $p_{\mathrm{int}}^t$ from $z_t^{obs}$, supervised by the trajectory label $o_t^{\mathrm{traj}}$ defined in Sec. 4.1 with a cross-entropy loss $\mathcal{L}_{\mathrm{traj}}$. This stage is optimized with the following multi-task objective:
\begin{equation}
	\mathcal{L}_{\mathrm{stage1}} = \lambda_v \mathcal{L}_{\mathrm{vae}} + \lambda_f \mathcal{L}_{\mathrm{for}} + \lambda_c \mathcal{L}_{\mathrm{inv}} + \lambda_t \mathcal{L}_{\mathrm{traj}}.
\end{equation}
After training, the observation encoder $E_\theta$ maps hand-object geometric observations into latent representations that capture the coupling between action-driven state transitions and proximity changes. These representations provide the state basis for downstream policy generation.

\subsection{Trajectory-Adaptive Proximity Policy}
\label{sec:trajectory_adaptive_policy}

During policy learning, we freeze the observation encoder trained in the previous stage and use a Transformer-based denoising policy to generate actions. The core design is a trajectory-adaptive asymmetric causal encoder. Specifically, the global observation feature $z_t^{obs}$ is fully visible to all action steps and provides high-level guidance about the manipulation scene. In contrast, the proximity-token sequence $z_t^{\mathrm{prox}}$ is constrained by a causal memory mask $M_{\mathrm{causal}}$, so that the action at step $s$ can access only the current and historical proximity states (Fig.~\ref{fig2}(c)). This branch is therefore responsible for local action refinement over time. Architectural details are given in Appendix~\ref{app:trajectory_adaptive_policy}. During Stage-2 training, $z_t^{\mathrm{prox}}$ is obtained from the frozen VAE, whereas at inference it is replaced by the IDM estimate $\hat{z}_t^{\mathrm{prox}}$, which is trained to match the same VAE latent space through $\mathcal{L}_{\mathrm{inv}}$.

We further use the interaction-phase probability $p_t^{\mathrm{int}}$ from the trajectory head in Sec.~\ref{sec:latent_interaction_dynamics} to adaptively regulate the proximity branch: the local interaction branch is suppressed during the motion phase and strengthened during the interaction phase. Given an expert action sequence $A_0$, diffusion step $k$, noisy action $A_k$, and noise $\epsilon$, the training objective of the second stage is:
\begin{equation}
	\mathcal{L}_{\mathrm{stage2}}
	=
	\mathbb{E}_{A_0,\epsilon,k}
	\left[
	\left\|
	\epsilon-
	\epsilon_\theta
	\left(
	A_k,k
	\mid
	z_t^{obs},\,
	g(p_t^{\mathrm{int}})z_t^{\mathrm{prox}}
	\right)
	\right\|_2^2
	\right]
	+
	\lambda_{\mathrm{gate}}\mathcal{L}_{\mathrm{gate}}
\end{equation}
Here, $g(p_t^{\mathrm{int}})$ is a gate conditioned on the interaction-phase probability, and $\mathcal{L}_{\mathrm{gate}}$ regularizes the gate distribution. This design allows the policy to rely on global visual guidance during the motion phase and adaptively strengthen proximity feedback during the interaction phase, producing more stable contact-aware actions throughout dexterous manipulation.

\begin{figure}[!t]
	\centering
	\captionsetup{skip=4pt}
	\includegraphics[width=1.0\textwidth]{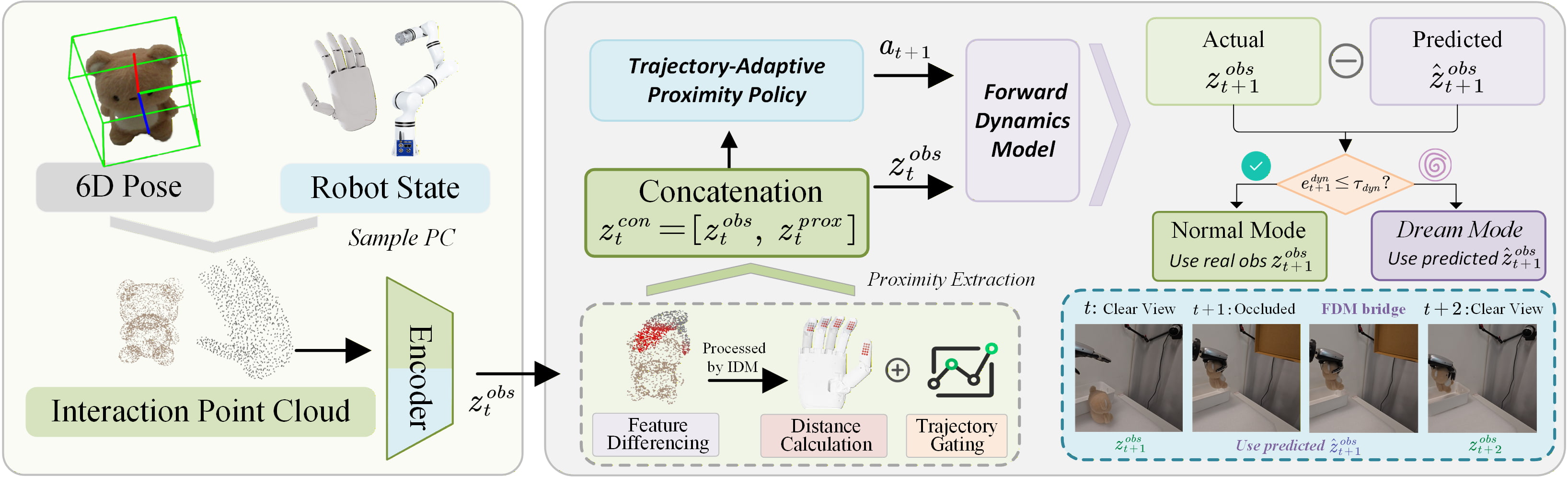}
	\caption{Overview of the Policy Inference Pipeline with Dynamics Consistency Guidance.
	}
	\label{fig3}
	\vspace{-10pt}
\end{figure}

\subsection{Dynamics-Consistency Guided Policy Inference}
\label{sec:dynamics_guided_inference}

During online inference, we combine the policy with the forward dynamics model $\mathrm{FDM}(\cdot)$ to form a closed-loop controller with physical consistency checking, as shown in Fig.~\ref{fig3}. At each time step, the observation encoder $E_\theta$ extracts the latent variable $z_t^{obs}$ from the real-time interaction point cloud $o_t^p$ and robot state $s_t$, and estimates the current proximity latent feature $\hat{z}_t^{\mathrm{prox}}$ through the $\mathrm{IDM}(\cdot)$. The trajectory head then outputs the interaction probability $p_t^{\mathrm{int}}$, which adaptively modulates the proximity branch. The policy generates an action sequence conditioned on $z_t^{obs}$, $\hat{z}_t^{\mathrm{prox}}$, and $p_t^{\mathrm{int}}$, and executes it in a receding-horizon manner.

To handle occlusion or short-term perception failure, the system uses $\mathrm{FDM}(\cdot)$ to predict the next latent variable and computes the dynamics deviation between the prediction and the real observation:
\begin{equation}
	e^{\mathrm{dyn}}_{t+1}=\left\|z_{t+1}^{obs}-\mathrm{FDM}(z_t^{obs},a_t)\right\|_2
\end{equation}
When $e_{t+1}^{\mathrm{dyn}}\leq\tau_{\mathrm{dyn}}$, ProxiDex regards real-time visual feedback as reliable and updates the policy state with the perceived latent observation. Otherwise, it switches to dynamics-dream mode, where action generation is maintained using latent variables autoregressively predicted by $\mathrm{FDM}(\cdot)$. The system returns to regular closed-loop inference once the deviation $e_{t+1}^{\mathrm{dyn}}$ falls below the threshold $\tau_{\mathrm{dyn}}$. In deployment, FoundationPose++ runs at around $10$ Hz, the proximity policy at about $28$ Hz, and $\mathrm{FDM}(\cdot)$ at about $20$ Hz, which is sufficient for receding-horizon inference; implementation details are given in Appendix~\ref{app:closed_loop_inference}.

\begin{table}[!t]
	\centering
	\caption{Simulation results on dexterous manipulation benchmarks. We evaluate all methods on Adroit, DexArt, and three self-designed IsaacLab tasks.}
	\label{tab:algorithm_performance} 
	\small 
	\renewcommand{\arraystretch}{1.2}
	\setlength{\tabcolsep}{4pt} 
	\begin{tabular}{c|c|c|cccc}
		\specialrule{1.2pt}{0pt}{0pt} 
		Algorithm\textbackslash{}Task & Obs. & Pretrain & Adroit 3 Tasks & DexArt 4 Tasks & Self-Designed 3 Tasks & Avg. \\ \hline
		DP3 \cite{ze2024dp3} & PC     & \textcolor{red}{\(\times\)} & 77.8\(\pm\)2.4 & 60.6\(\pm\)0.7 & 80.8\(\pm\)2.3 & 71.8\(\pm\)1.7 \\ 
		ManiFlow \cite{yan2025maniflow} & PC     & \textcolor{red}{\(\times\)} & 78.6\(\pm\)2.3 & 63.3\(\pm\)2.7 & 84.4\(\pm\)1.9 & 74.2\(\pm\)2.3 \\ 
		AFRO \cite{liang2025bootstrap} & PC     & \textcolor{green}{\(\checkmark\)} & 84.0\(\pm\)2.8 & 64.7\(\pm\)2.5 & 86.0\(\pm\)1.5 & 76.9\(\pm\)2.3 \\ 
		CordViP \cite{fu2025cordvip}& Int.PC & \textcolor{red}{\(\times\)} & 85.6\(\pm\)2.3 & 71.8\(\pm\)1.8 & 88.8\(\pm\)1.6 & 81.0\(\pm\)1.9 \\ 
		\textbf{ProxiDex (Ours)} & \multicolumn{1}{c|}{Int.PC} & \textcolor{green}{\(\checkmark\)} &
		\cellcolor{gray!20}\textbf{90.3\(\pm\)1.4} & \cellcolor{gray!20}\textbf{73.7\(\pm\)2.2} & \cellcolor{gray!20}\textbf{91.1\(\pm\)1.7} & \cellcolor{gray!20}\textbf{83.9\(\pm\)1.8} \\ 
		\specialrule{1.2pt}{0pt}{0pt} 
	\end{tabular}
	\vspace{-10pt}
\end{table}

\section{Experiments}
We evaluate ProxiDex in both simulation and real-world settings, focusing on two aspects: (i) whether interaction point clouds and proximity cues can improve dexterous manipulation performance; (ii) whether the learned interaction dynamics can maintain stable control under disturbances.

\subsection{Simulation Experiments}
\textbf{Setup.} We evaluate ProxiDex on $10$ multi-finger tasks from Adroit \cite{rajeswaran2017learning}, DexArt \cite{bao2023dexart}, and our customized IsaacLab environments \cite{mittal2025isaac}. Demonstrations for Adroit and DexArt are generated by reinforcement-learning experts, while customized tasks are collected through VR teleoperation. For each task, we collect $30$ high-quality demonstrations covering representative hand--object contact patterns. Since object poses are directly available in simulation, we can accurately generate interaction point clouds $o_t^p$ and region-level proximity features $o_t^{\mathrm{prox}}$, as visualized in Appendix~Fig.~\ref{fig:simulation_point_clouds}. Further details are provided in Appendices~\ref{app:simulation_tasks} and~\ref{app:simulation_details}.

\textbf{Baselines.} We compare ProxiDex with four representative 3D point cloud policies. DP3 \cite{ze2024dp3} and ManiFlow \cite{yan2025maniflow} use camera scene point clouds (PC) for end-to-end action prediction. AFRO \cite{liang2025bootstrap} uses the same input with dynamics pretraining to evaluate the effect of latent dynamics. CordViP \cite{fu2025cordvip} adopts reconstructed interaction point clouds (Int.PC) and an arm-coordinated denoising policy.

\textbf{Results.} Table~\ref{tab:algorithm_performance} summarizes the benchmark-level performance averaged over the evaluated tasks. ProxiDex achieves the best performance on all three task groups, reaching an overall success rate of $83.9\%$ and outperforming DP3, ManiFlow, AFRO, and CordViP by $12.1$, $9.7$, $7.0$, and $2.9$ percentage points, respectively. On Adroit, DexArt, and our customized IsaacLab tasks, ProxiDex obtains $90.3\%$, $73.7\%$, and $91.1\%$ success rates, exceeding CordViP by $4.7$, $1.9$, and $2.3$ percentage points. Since CordViP also uses interaction point clouds, these gains mainly come from modeling fine-grained hand--object proximity and its temporal evolution. This is especially beneficial for contact-sensitive tasks such as Pen and Driller, where the policy must continuously adjust fingertip placement and wrist pose after contact under occlusion and object motion.

\begin{wrapfigure}{r}{0.4\textwidth}
	\vspace{-8pt}
	\centering
	\includegraphics[width=0.4\textwidth]{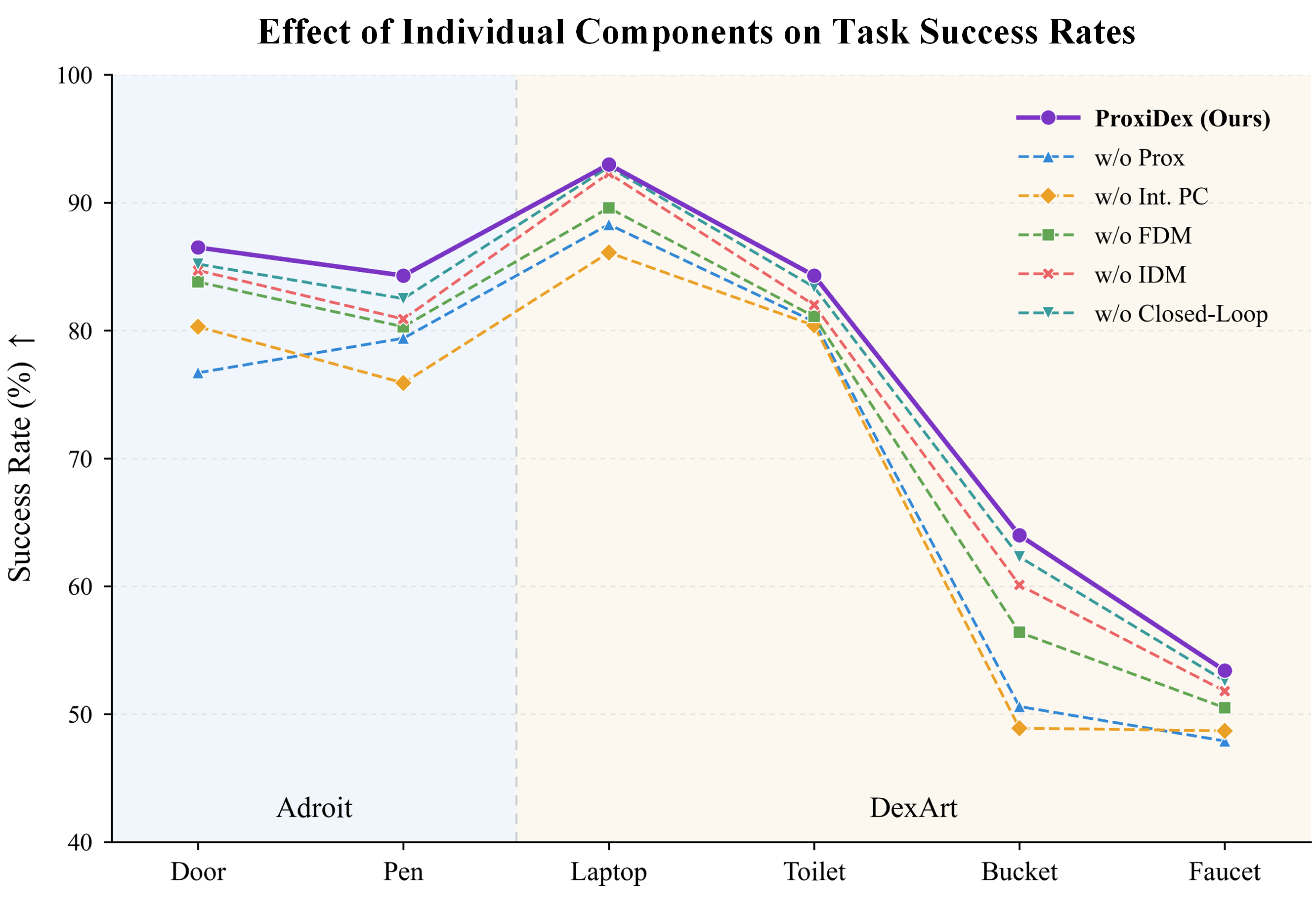}
	\caption{Ablation on Success Rate.}
	\label{fig4_55}
	\vspace{-10pt}
\end{wrapfigure}

\textbf{Ablation on Task Success Rate.} To further analyze the contribution of each core component to task success, Fig.~\ref{fig4_55} presents task-wise ablations on six representative Adroit and DexArt tasks. ProxiDex performs best across all tasks, showing the joint benefit of interaction-centric representation and dynamics-aware policy learning. Removing proximity cues or replacing interaction point clouds with camera-observed scene point clouds causes clear drops, especially on contact-sensitive tasks such as Bucket and Pen, highlighting the importance of explicit hand--object interaction geometry under occlusion. Removing FDM, IDM, or closed-loop dynamics checking also reduces task success, indicating that latent dynamics modeling and online verification help improve execution stability during contact-rich manipulation.

\begin{wrapfigure}{r}{0.43\textwidth}
	\vspace{-8pt}
	\centering
	\includegraphics[width=0.4\textwidth]{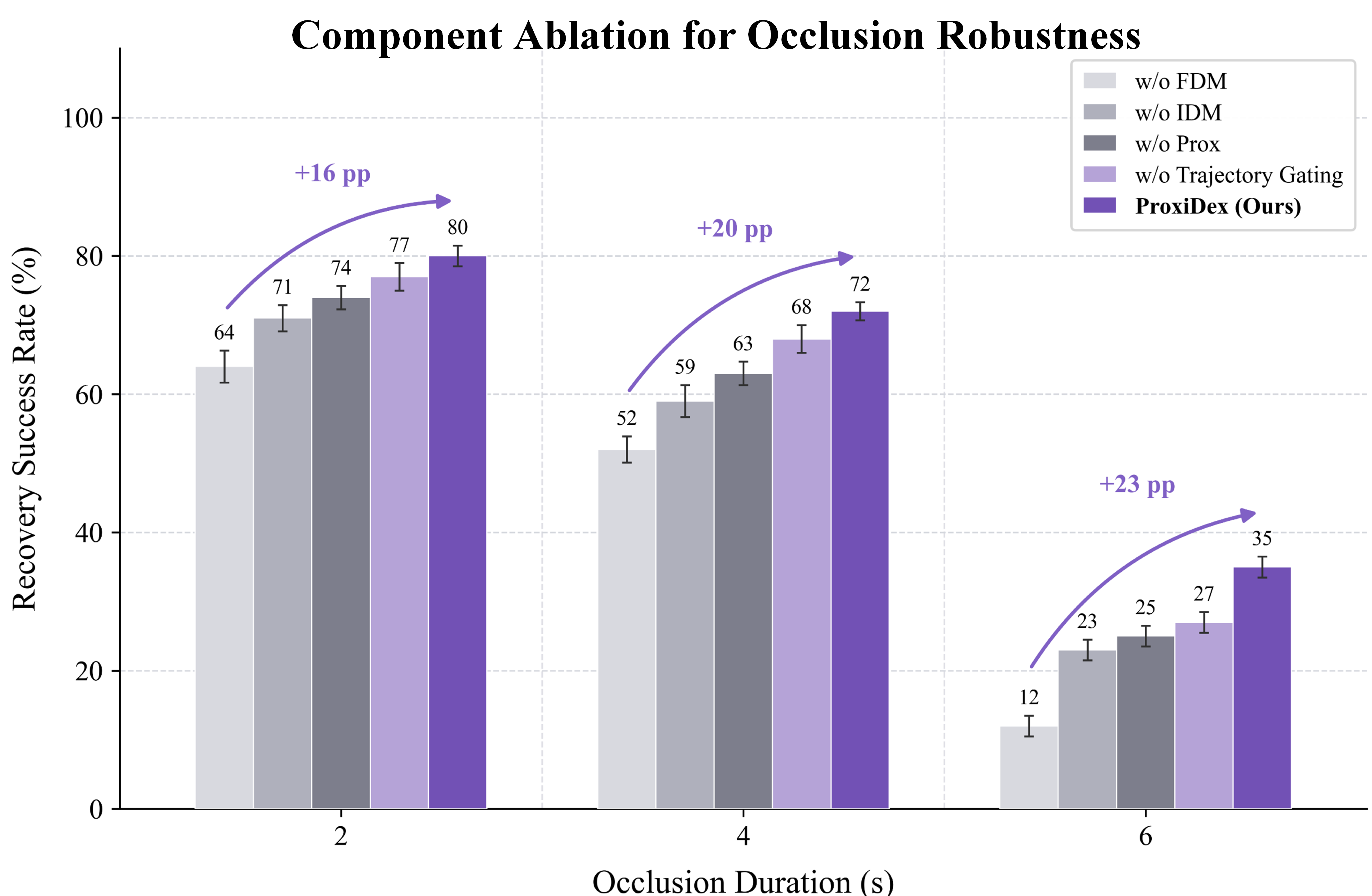}
	\caption{Ablation on Occlusion.}
	\label{fig5}
	\vspace{-10pt}
\end{wrapfigure}

\textbf{Ablation on Occlusion Robustness.} To evaluate component contributions to occlusion robustness, we introduce a cuboid occluder into the depth-camera view for 2, 4, and 6 s, and measure recovery success on the Adroit Pen task. As shown in Fig.~\ref{fig5}, full ProxiDex achieves 80\% and 72\% recovery under 2-s and 4-s occlusions. Removing FDM causes the largest drop, as future-latent prediction and dynamics takeover are no longer available. Without IDM, hand--object proximity is harder to recover from latent transitions, reducing success to 71\% and 59\%. Removing proximity cues or trajectory gating also weakens local interaction modeling and phase-adaptive modulation. Under 6-s occlusion, accumulated prediction errors reduce the full model to 35\% recovery.

\subsection{Real-World Experiments}
\begin{figure}[!t]
	\centering
	\captionsetup{skip=4pt}
	\includegraphics[width=1.0\textwidth]{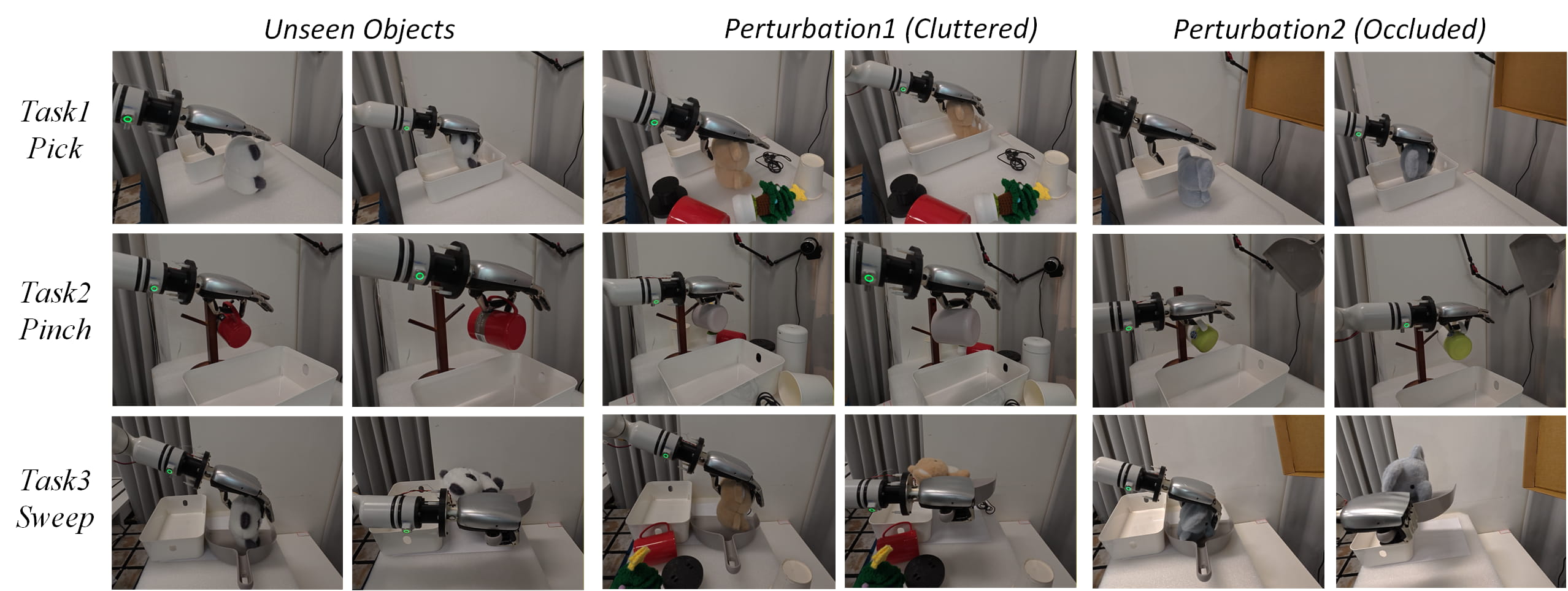}
	\caption{We introduce unseen objects, cluttered scenes, and visual occlusions into three tabletop manipulation tasks. The results show that ProxiDex can maintain stable hand-object interaction and complete the corresponding manipulation tasks in complex real-world scenarios.}
	\label{fig4}
	\vspace{-10pt}
\end{figure}

\textbf{Setup.} The real-robot platform consists of a Realman RM75B 7-DoF robotic arm, a RealSense L515 RGB-D camera, and two dexterous hands. Specifically, a CasBot-P1L hand with 6 active DoFs is used for the standard tabletop manipulation tasks, while a 20-DoF Wuji Hand V1 is used for the more contact-rich manipulation tasks. As shown in Fig.~\ref{fig4} and Fig.~\ref{fig:real_execution_point_clouds}, we design three tabletop tasks. For each task, we collect $50$ demonstrations and evaluate the policy under in-distribution objects, unseen objects, and disturbance conditions. More details are provided in Appendix~\ref{app:real_world}.

\textbf{Results on Tabletop Tasks.} Table~\ref{tab1:task_success} shows that ProxiDex achieves the highest real-robot success rate, with an overall average of $72.6\%$. It outperforms CordViP, AFRO, ManiFlow, and DP3 by $9.7$, $19.1$, $22.9$, and $32.6$ percentage points, respectively. ProxiDex also obtains the best or tied-best result in nine task settings, demonstrating that the proposed interaction-centric representation and proximity-aware policy consistently improve real-world dexterous manipulation performance.

\begin{table}[!t]
	\centering
	\caption{Real-robot task success counts under in-distribution, unseen-object, perturbation, and contact-rich settings. The results show that ProxiDex achieves accurate and robust performance.}
	\label{tab1:task_success}
	\small
	\setlength{\tabcolsep}{3.2pt}
	\renewcommand{\arraystretch}{1.2}
	
	\begin{tabular}{@{}c|ccc|ccc|ccc|cc|c@{}}
		\toprule
		
		\multirow{2}{*}[-0.5ex]{\shortstack[c]{Tasks\\[2pt]\\Algorithm}}
		& \multicolumn{3}{c|}{\textbf{In Distribution}}
		& \multicolumn{3}{c|}{\textbf{Unseen Objects}}
		& \multicolumn{3}{c|}{\textbf{Perturbation}}
		& \multicolumn{2}{c|}{\textbf{Contact-Rich}}
		& \multirow{2}{*}{\textbf{Avg.}} \\
		
		\cmidrule(lr){2-4}
		\cmidrule(lr){5-7}
		\cmidrule(lr){8-10}
		\cmidrule(lr){11-12}
		
		& Pick & Pinch & Sweep
		& Pick & Pinch & Sweep
		& Pick & Pinch & Sweep
		& Twist & Flip
		& \\
		
		\midrule
		
		DP3 \cite{ze2024dp3}
		& 14/20 & 10/20 & 12/20
		& 18/40 & 9/40  & 16/40
		& 14/30 & 13/30 & 14/30
		& 2/20  & 2/20
		& 40.0\% \\
		
		ManiFlow \cite{yan2025maniflow}
		& 16/20 & 11/20 & 15/20
		& 24/40 & 12/40 & 19/40
		& 17/30 & 18/30 & 17/30
		& 2/20  & 3/20
		& 49.7\% \\
		
		AFRO \cite{liang2025bootstrap}
		& 17/20 & 11/20 & 16/20
		& 25/40 & 15/40 & 24/40
		& 17/30 & 17/30 & 16/30
		& 3/20  & 5/20
		& 53.5\% \\
		
		CordViP \cite{fu2025cordvip}
		& 18/20 & 13/20 & 17/20
		& 29/40 & 21/40 & 27/40
		& 19/30 & 19/30 & 18/30
		& 6/20  & 8/20
		& 62.9\% \\
		
		\midrule
		
		ProxiDex-1
		& 18/20 & 14/20 & 17/20
		& 30/40 & 22/40 & 28/40
		& 20/30 & 20/30 & 20/30
		& 7/20  & 6/20
		& 65.2\% \\
		
		ProxiDex-2
		& 18/20 & 14/20 & 18/20
		& 31/40 & 23/40 & 28/40
		& 21/30 & 22/30 & 21/30
		& 8/20  & 7/20
		& 68.1\% \\
		
		\textbf{ProxiDex}
		&
		\cellcolor{gray!20}\textbf{19/20} &
		\cellcolor{gray!20}\textbf{15/20} &
		\cellcolor{gray!20}\textbf{18/20} &
		\cellcolor{gray!20}\textbf{32/40} &
		\cellcolor{gray!20}\textbf{24/40} &
		\cellcolor{gray!20}\textbf{29/40} &
		\cellcolor{gray!20}\textbf{22/30} &
		\cellcolor{gray!20}\textbf{23/30} &
		\cellcolor{gray!20}\textbf{22/30} &
		\cellcolor{gray!20}\textbf{10/20} &
		\cellcolor{gray!20}\textbf{11/20} &
		\cellcolor{gray!20}\textbf{72.6\%} \\
		
		\bottomrule
	\end{tabular}
	
	\vspace{-10pt}
\end{table}

\textbf{Generalization and robustness.} Beyond the overall performance, Table 2 further shows that ProxiDex remains effective under distribution shifts and external disturbances. For unseen objects, it succeeds in $85/120$ trials, compared with $77/120$ for CordViP. Under perturbations, ProxiDex achieves $67/90$ successful trials, while CordViP obtains $56/90$.  \begin{wrapfigure}{r}{0.4\textwidth}
	\vspace{-8pt}
	\centering
	\includegraphics[width=0.38\textwidth]{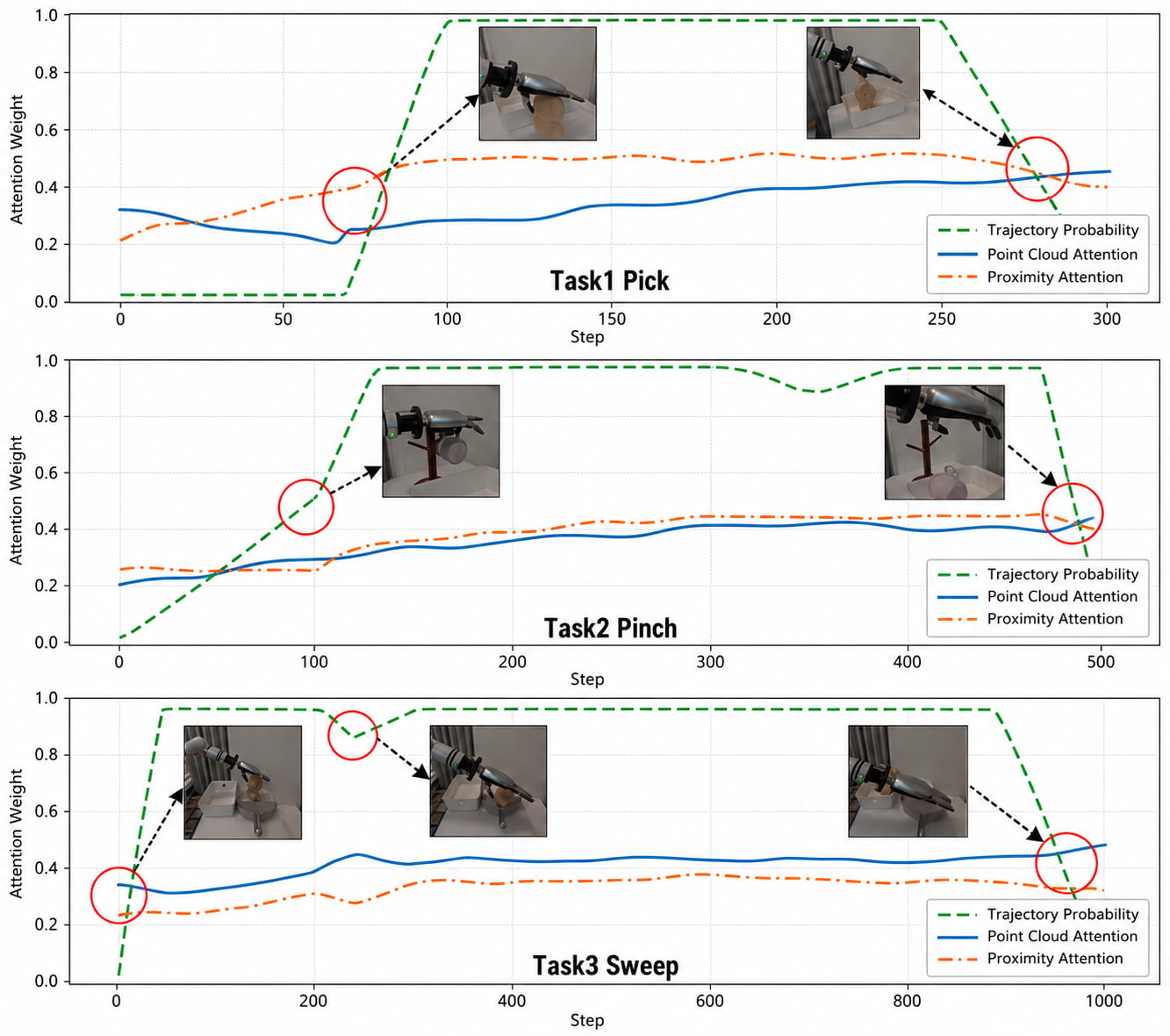}
	\caption{Attention visualization. Averaged Transformer attention weights of point-cloud and proximity tokens across three tabletop tasks.}
	\label{fig4_555}
	\vspace{-10pt}
\end{wrapfigure} These results indicate that explicit hand--object proximity modeling improves robustness to object variations and external disturbances.  In contrast, visible-scene point-cloud methods such as DP3 and ManiFlow degrade more clearly in Sweep, where hand occlusion and object pose changes make the observed scene geometry unstable.

\textbf{Backbone Ablation.} To further examine whether the gains only come from the policy backbone, the additional rows in Table~\ref{tab1:task_success} compare different policy backbones within ProxiDex. ProxiDex-1 with a U-Net backbone and ProxiDex-2 with a standard Transformer achieve $65.2\%$ and $68.1\%$ average success rates. The full ProxiDex further improves them by $7.4$ and $4.5$ percentage points, showing that the proposed backbone can emphasize hand--object proximity at critical contact moments. As shown in Fig.~\ref{fig4_555}, the interaction-phase probability is used to adaptively adjust the attention weights between proximity and point-cloud tokens, allowing the policy to emphasize hand-object proximity at critical interaction moments. 

\begin{wrapfigure}{r}{0.43\textwidth}
	\vspace{-8pt}
	\centering
	\includegraphics[width=0.4\textwidth]{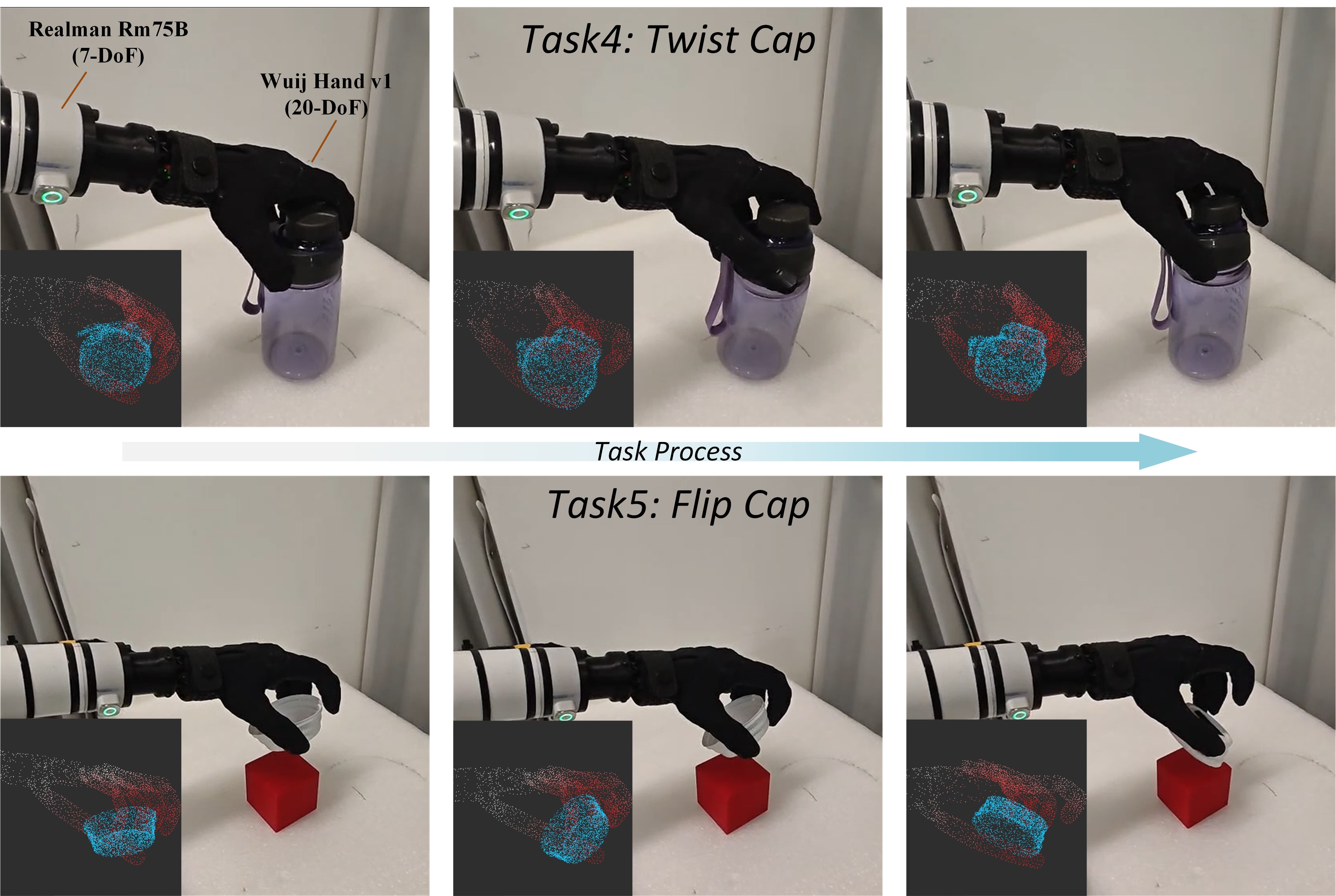}
	\caption{ProxiDex performance and interaction point-cloud visualization on contact-rich tasks.}
	\label{fig8}
	\vspace{-10pt}
\end{wrapfigure}

\textbf{Contact-Rich Task Analysis.}
We further evaluate two contact-rich tasks involving sustained finger--object interaction and object reorientation. As shown in Fig.~\ref{fig8}, ProxiDex maintains stable interaction representations and completes complex in-hand manipulation even under short-term visual occlusion. The interaction point cloud provides local hand--object geometry, while dynamics prediction preserves interaction-state continuity when visual feedback becomes temporarily unreliable. As shown in Table~\ref{tab1:task_success}, ProxiDex achieves the best performance on both contact-rich tasks, indicating that their combination improves stability during sustained contact and object reorientation.

\begin{wrapfigure}{r}{0.43\textwidth}
	\vspace{-8pt}
	\centering
	\includegraphics[width=0.4\textwidth]{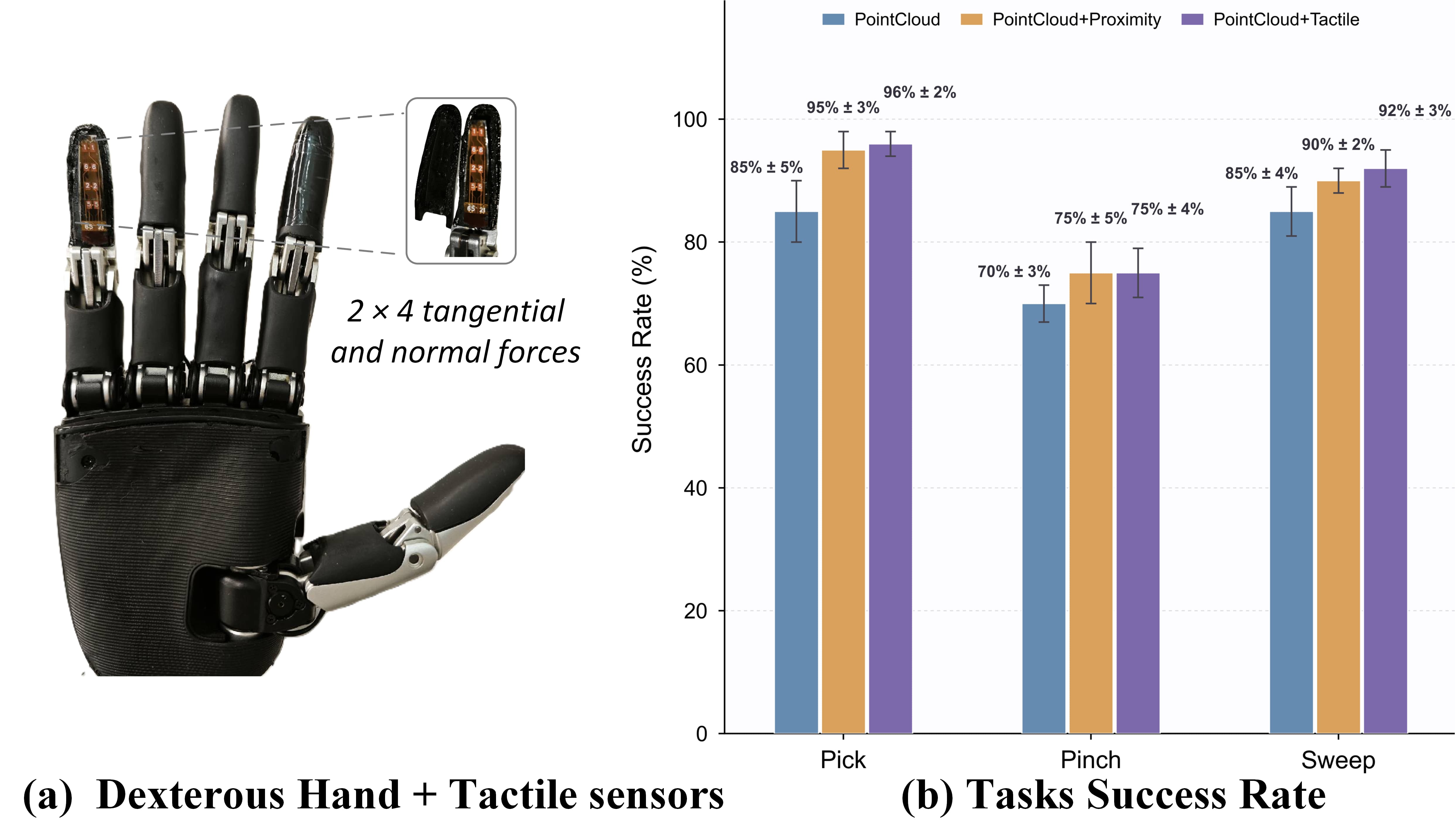}
	\caption{(a) P1L hand with tactile arrays. (b) Proximity--tactile comparison.}
	\label{fig9}
	\vspace{-10pt}
\end{wrapfigure}

\textbf{Comparison with Real Tactile Sensing.}
To compare proximity with real tactile feedback, each fingertip of the P1L hand is equipped with a $2\times4$ tactile array. As shown in Fig.~\ref{fig9}, proximity achieves competitive task performance compared with tactile sensing. Although proximity cannot replace force feedback, it captures changes in hand--object distance and spatial distribution before and around contact, providing useful geometric cues for pre-grasp adjustment and subsequent contact establishment.

\section{Conclusion}
We presented ProxiDex, a proximity-policy learning framework for dexterous manipulation. ProxiDex constructs region-level proximity representations from reconstructed hand--object interaction point clouds and further learns action-conditioned latent interaction dynamics to guide trajectory-adaptive policy learning and closed-loop inference. By jointly modeling local hand--object geometry and its dynamic evolution, ProxiDex supports stable policy execution under short-term unreliable visual feedback, sustained contact, and complex object interactions. Simulation and real-world experiments show consistent performance across standard, unseen-object, disturbance, visual-occlusion, and contact-rich settings. Overall, these results demonstrate that ProxiDex provides a robust and physically meaningful interaction modeling framework for dexterous manipulation.

\section{Limitations}
Although ProxiDex shows strong stability, it still has several limitations. First, it depends on accurate object reconstruction and real-time pose tracking, which may be unreliable for small components, articulated objects, or high-precision manipulation. Second, dynamics takeover is mainly effective for short-term perception failures, while prolonged occlusion can accumulate autoregressive prediction errors and reduce control reliability. Finally, the current proximity representation remains geometry-driven and cannot measure real contact forces or precise fingertip force distributions. Future work will combine improved pose estimation, real tactile sensing, and large-scale human manipulation videos to enhance long-horizon recovery and fine-grained contact modeling.


\clearpage
\acknowledgments{This work was supported by the Hebei Provincial Science and Technology Plan Project under Grant No. 25241802D, the National Natural Science Foundation of China under Grant No. 62303457,  and the New Generation Artificial Intelligence-National Science and Technology Major Project under Grant No. 2025ZD0122900.}


\bibliography{ProxiDex}  

\clearpage
\appendix

\begin{center}
	{\Large \bfseries Supplementary Material for ProxiDex}
\end{center}
\vspace{0.5em}

This supplementary material provides additional details on data acquisition, latent dynamics modeling, policy learning, and experimental protocols. It is organized as follows:
\begin{itemize}
	\setlength{\itemsep}{2pt}
	\setlength{\parsep}{0pt}
	\item Appendix~\ref{app:data_representation} Proximity-guided data acquisition and geometric representation.
	\item Appendix~\ref{app:training_inference} Training and closed-loop inference details of ProxiDex.
	\item Appendix~\ref{app:simulation_setup} Simulation setup and qualitative analysis.
	\item Appendix~\ref{app:additional_analysis} Additional experiments and ablation studies.
	\item Appendix~\ref{app:real_world} Real-world experiment setup and evaluation protocol.
\end{itemize}

\section{Proximity-Guided Data Acquisition and Geometric Representation}
\label{app:data_representation}

\subsection{VR Teleoperation and Action Retargeting}
\label{app:vr_teleoperation}

\textbf{Arm teleoperation.} We use incremental inverse kinematics (IK) to retarget VR wrist motion to the robot end-effector. The system \cite{weng2026hts} first records the initial VR wrist position $\boldsymbol{x}_{\rm init}$ and the initial robot end-effector position $\boldsymbol{x}_{\rm home}$. At runtime, it computes the displacement of the current VR wrist position $\boldsymbol{x}_{\rm vr}$ relative to the calibrated origin, and maps this displacement into the robot base frame using a scale vector $\boldsymbol{S}$ and a coordinate retargeting matrix $\boldsymbol{R}_{\rm map}$:
\begin{equation}
	\boldsymbol{x}_{\rm target}=\boldsymbol{x}_{\rm home}+\mathbf{R}_{\rm map}\left(\mathbf{S}\circ(\boldsymbol{x}_{\rm vr}-\boldsymbol{x}_{\rm init})\right),
\end{equation}
where $\mathbf{S}$ is an anisotropic scaling vector, $\mathbf{R}_{\rm map}$ aligns the VR coordinate frame with the robot base frame, and $\circ$ denotes element-wise multiplication. The target pose is solved by numerical inverse kinematics under joint-limit and redundancy constraints, and the resulting joint commands are sent to the Realman RM75B arm through the low-level controller.

For orientation control, the system uses quaternion increments to synchronize the VR wrist orientation and the robot end-effector orientation. Let $q_{\rm vr}$ be the current VR wrist orientation, $q_{\rm init}$ the initial calibrated orientation, and $q_{\rm home}$ the initial robot end-effector orientation. The target orientation is
\begin{equation}
	\Delta q=q_{\rm vr}\otimes q_{\rm init}^{-1},
	\qquad
	q_{\rm target}=q_{\rm home}\otimes \Delta q .
\end{equation}
The Cartesian target is passed to a numerical IK solver in PyBullet, which computes joint commands under 7-DoF redundancy and joint-limit constraints. The resulting commands are synchronized to the RealMan RM75B arm through CAN-FD, enabling low-latency spatial following.

\textbf{Dexterous-hand teleoperation.} The VR detector provides $21$ hand keypoints $\boldsymbol{u}_i$ \cite{weng2026hts}. We normalize them by removing the wrist translation and applying a scale factor, i.e., $\tilde{\boldsymbol{u}}_i=\eta(\boldsymbol{u}_i-\boldsymbol{u}_0)$, followed by a handedness correction when needed. A local wrist frame is estimated from the wrist and finger-root keypoints, and the normalized keypoints are reprojected into a canonical hand coordinate system. This step reduces differences among operators and provides a consistent geometric input for retargeting.

The retargeting problem converts human-hand topology into robot joint commands. We construct target vector flows $\hat{v}_k$ that describe the relative geometry of the human fingers and solve for the robot joint configuration $q_t^*$ by minimizing
\begin{equation}
	q_t^* = \arg\min_q \sum_k w_k \rho_\delta\left(\|r_k(q)-\hat{v}_k\|_2^2\right)+\lambda\|q-q_{t-1}\|_2^2 .
\end{equation}
Here, $\rho_\delta$ is the Huber loss for robustness to keypoint noise, and the second term regularizes temporal smoothness by penalizing abrupt joint changes. We further introduce a DexPilot-style grasp prior projection \cite{handa2020dexpilot} and update the projection state with threshold hysteresis, improving fingertip closure while reducing hand jitter.

\begin{figure}[!t]
	\centering
	\captionsetup{skip=4pt}
	\includegraphics[width=1.0\textwidth]{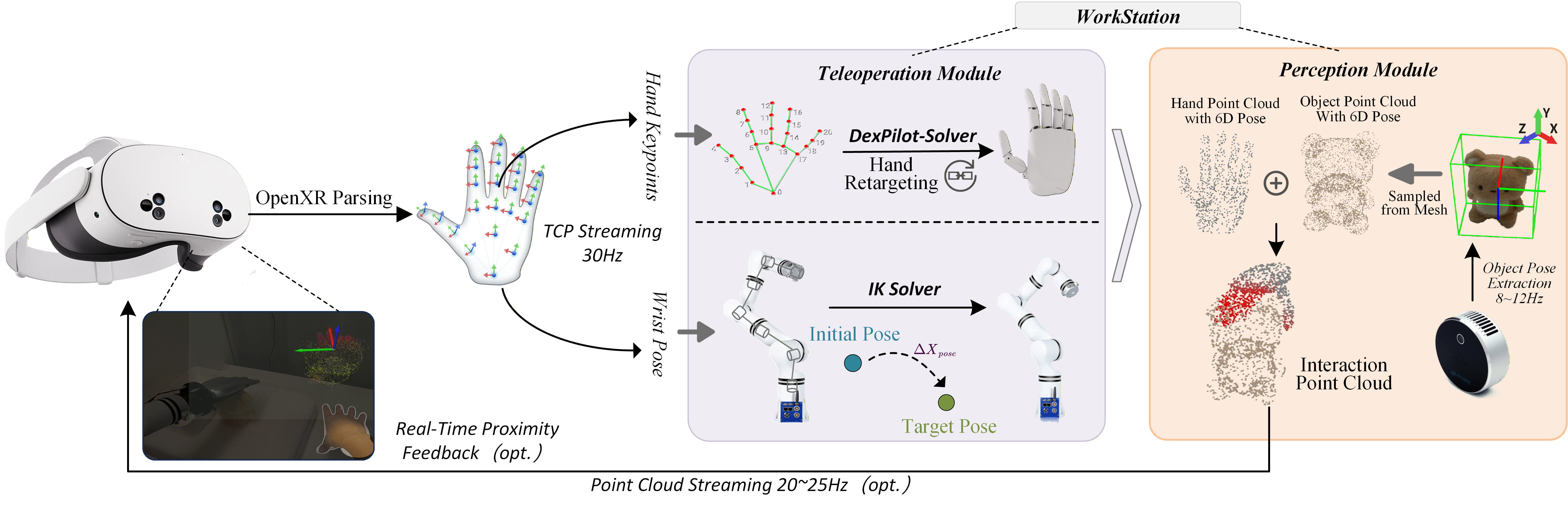}
	\caption{VR teleoperation pipeline in ProxiDex, where OpenXR hand keypoints and wrist poses are streamed to a workstation for hand--arm control, and the reconstructed hand--object interaction point cloud is sent back to the headset as optional proximity feedback for immersive teleoperation.}
	\label{fig:vr_teleoperation}
	\vspace{-10pt}
\end{figure}

The teleoperation loop is sufficiently responsive for real-time operation. TCP hand-joint commands are transmitted at approximately $30$ Hz, object-pose detection runs at around $10$ Hz, and the interaction point-cloud stream is updated at about $23$ Hz. These rates allow the operator to receive timely visual and proximity feedback during manipulation, as summarized in Fig.~\ref{fig:vr_teleoperation}. In particular, when direct visual observation is degraded by hand-object occlusions, the proximity feedback provides an augmented-reality style cue that helps the operator perceive the hand-object distance and adjust the manipulation process more intuitively. With the teleoperation stream defined, the next step is to convert the reconstructed object geometry into a metric-scale representation for distance-based proximity computation.

\subsection{Object Reconstruction and Metric-Scale Calibration}
\label{app:object_reconstruction}

We further detail how ProxiDex reconstructs occlusion-free hand-object interaction point clouds from demonstration data. We first use Grounded-SAM\cite{ren2024grounded} to segment the target object from the initial RGB observation, and then reconstruct the target mesh $\mathcal{M}_o$ using SAM3D\cite{chen2025sam}. Monocular reconstruction can complete invisible regions and provide a full geometric prior, but its output usually lacks reliable metric scale. Therefore, before using the mesh for interaction point-cloud synthesis and proximity computation, we perform scale calibration through the following procedure.

For each object, the system captures one aligned RealSense RGB-D observation and estimates a scale factor $\kappa$ from the depth map. Given an RGB image, a metric depth map, and camera intrinsics, Grounded-SAM first produces the target object mask. Valid depth pixels inside the mask are then back-projected into the camera frame:
\begin{equation}
	p(u,v)=\left(\frac{(u-c_x)d}{f_x},\frac{(v-c_y)d}{f_y},d\right),
\end{equation}
where $(u,v)$ denotes pixel coordinates, $d$ is the depth value, and $(c_x,c_y)$ and $(f_x,f_y)$ are the principal point and focal lengths. After removing statistical outliers from the back-projected object point cloud, we compute its oriented bounding-box size $\boldsymbol{e}_{\rm rgbd}$.

We also load the vertices of the OBJ mesh reconstructed by SAM3D and compute its oriented bounding-box size $\boldsymbol{e}_{\rm obj}$. To avoid axis-order effects in scale estimation, both size vectors are sorted before comparison:
\begin{equation}
	\boldsymbol{r}=\frac{\mathrm{sort}(e_{\rm rgbd})}{\mathrm{sort}(e_{\rm obj})} .
\end{equation}
The final uniform scale coefficient is $\kappa=\mathrm{median}(\boldsymbol{r})$, which is applied to the mesh vertices. This calibration preserves the shape-completion advantage of SAM3D while grounding subsequent distance calculations in metric geometry. Using the median of the sorted axis ratios also reduces sensitivity to a single noisy bounding-box dimension, which can appear when the RGB-D mask contains small boundary artifacts. The calibrated mesh therefore provides a stable object surface for the region-level proximity computation described next.

\begin{table*}[!t]
	\centering
	\caption{Input data specifications across simulation and real-world environments.}
	\label{tab:input_data_specifications}
	\small
	\renewcommand{\arraystretch}{1.15}
	\resizebox{\textwidth}{!}{
		\begin{tabular}{llcccc}
			\toprule
			\textbf{Environment} & \textbf{Input Setting} & \textbf{Point-Cloud Input} & \textbf{Auxiliary Input} & \textbf{Action Dim.} & \textbf{Horizon / Notes} \\
			\midrule
			
			\multicolumn{6}{l}{\textit{Simulation Environments}} \\
			Adroit 
			& Scene Point Cloud 
			& $[512, 3]$ \texttt{point\_cloud} 
			& $[24]$ \texttt{low\_dim} state 
			& $28$ 
			& $n_{\mathrm{obs}}$ \\
			
			Adroit 
			& Interaction Point Cloud 
			& $[3 \times 512, 3]$ \texttt{point\_cloud} 
			& $[24]$ \texttt{low\_dim} state 
			& $28$ 
			& $N=3$, including hand and object \\
			
			\midrule
			DexArt 
			& Scene Point Cloud 
			& $[1024, 3]$ \texttt{point\_cloud} 
			& $[32]$ \texttt{low\_dim} state 
			& $22$ 
			& $n_{\mathrm{obs}}$ \\
			
			DexArt 
			& Interaction Point Cloud 
			& $[2 \times 1024, 3]$ \texttt{point\_cloud} 
			& $[32]$ \texttt{low\_dim} state 
			& $22$ 
			& $N=2$, including hand and object \\
			
			\midrule
			IsaacLab 
			& Scene Point Cloud 
			& $[1024, 3]$ \texttt{point\_cloud} 
			& $[18]$ \texttt{low\_dim} state 
			& $18$ 
			& $n_{\mathrm{obs}}$ \\
			
			IsaacLab 
			& Interaction Point Cloud 
			& $[N \times 1024, 3]$ \texttt{point\_cloud} 
			& $[18]$ \texttt{low\_dim} state 
			& $18$ 
			& $N=2$ for Box and $N=3$ for other tasks \\
			
			\midrule
			\multicolumn{6}{l}{\textit{Real-World Environment}} \\
			Real World 
			& Scene Point Cloud 
			& $[512, 3]$ \texttt{point\_cloud} 
			& $[13]$ \texttt{low\_dim} state 
			& $13$ 
			& $n_{\mathrm{obs}}$ \\
			
			Real World 
			& Interaction Point Cloud 
			& $[N \times 512, 3]$ \texttt{point\_cloud} 
			& $[13]$ \texttt{low\_dim} state 
			& $13$ 
			& $N=3$ for Sweep and $N=2$ for other tasks \\
			
			\bottomrule
		\end{tabular}
	}
	\vspace{-10pt}
\end{table*}

\subsection{Geometric Proximity Representation}
\label{app:geometric_proximity}

Given the metrically calibrated object mesh, the system samples points from the surfaces of object mesh and hand URDF links, and transforms them into the robot base frame using the object 6D pose tracked by FoundationPose++ \cite{Wenhao_Yan_and_Jie_Chu_FoundationPose_2025} and robot forward kinematics. This yields the object point cloud $P_t^o$ and the hand point cloud $P_t^h$. The interaction point cloud at frame $t$ is defined as $o_t^p=P_t^o\cup P_t^h$. Compared with a raw scene point cloud, this representation explicitly preserves the relative geometry between the hand and object while reducing the influence of background points and self-occlusion.

Proximity is computed from geometric distances between the hand and object point clouds, without assuming any specific tactile sensor. For each hand sample $p_i^h\in P_t^h$, we first compute its nearest-neighbor distance to the object point cloud:
\begin{equation}
	d_i^{\mathrm{nn}}=\min_{p_j^o\in P_t^o}\left\|p_i^h-p_j^o\right\|_2
\end{equation}
In practice, small geometric interpenetrations may occur due to pose-tracking noise, calibration errors, or simplified hand/object meshes. Since the proximity representation is intended to describe near-field interaction rather than penetration depth, we treat such cases as saturated contact responses. Specifically, using the signed distance field $\phi_{\mathcal{O}}(\cdot)$ of the object mesh, where negative values indicate points inside the object, the effective distance is clipped as
\begin{equation}
	d_i =
	\begin{cases}
		d_i^{\mathrm{nn}}, & \phi_{\mathcal{O}}(p_i^h)\geq 0,\\
		0, & \phi_{\mathcal{O}}(p_i^h)<0.
	\end{cases}
\end{equation}
This prevents occasional interpenetration artifacts from producing unstable proximity values while preserving consistent contact cues for policy learning. A Gaussian kernel then maps the effective distance to a normalized proximity response:
\begin{equation}
	\eta_i=\exp\left(-\frac{1}{2}\left(\frac{d_i}{\sigma}\right)^2\right),\qquad \eta_i\in[0,1],
\end{equation}
where $\sigma$ controls the decay sensitivity. For VR visualization, $\eta_i$ can be mapped to the hand point color $c_i=c_{\rm base}+\eta_i(c_{\rm hot}-c_{\rm base})$, providing proximity guidance related to object distance. For policy training, we record the full $o_t^p$ and proximity observation to maintain input consistency.

To obtain structured learning inputs, we group hand samples according to physical hand regions $S_k$, such as the palm, finger roots, and fingertips. Let $I_k=\{i\mid p_i^h\in S_k\}$ be the set of sample indices belonging to region $S_k$. The region-level proximity is
\begin{equation}
	\gamma_t^k=\frac{1}{|I_k|}\sum_{i\in I_k}\eta_i .
\end{equation}
The proximity observation at frame $t$ is therefore
\begin{equation}
	o_t^{\mathrm{prox}}=[\gamma_t^1,\dots,\gamma_t^K].
\end{equation}
We further derive the trajectory-stage label from the palm response:
\begin{equation}
	o_t^{\mathrm{traj}}=\mathbb{I}(\gamma_t^{\mathrm{palm}}>\tau_{\text{palm}}),
\end{equation}

where $o_t^{\mathrm{traj}}=0$ denotes the approach stage and $o_t^{\mathrm{traj}}=1$ denotes the contact-interaction stage. Each demonstration frame is finally stored as $(o_t^p,o_t^{\mathrm{prox}},o_t^{\mathrm{traj}},s_t,a_t)$, consistent with the notation in the main paper. The concrete input dimensions of scene point clouds, interaction point clouds, auxiliary states, and actions across simulation and real-world environments are summarized in Table~\ref{tab:input_data_specifications}. In this representation, $o_t^p$ keeps the complete hand--object geometry, $o_t^{\mathrm{prox}}$ compresses near-field distances into physically meaningful hand-region responses, and $o_t^{\mathrm{traj}}$ provides a coarse phase indicator. This separation is used in the following training stage, where the latent observation is explicitly aligned with proximity dynamics rather than only with visible geometry.

\section{Training and Inference Details of ProxiDex}
\label{app:training_inference}

ProxiDex is trained in two stages. The first stage learns a proximity-aware dynamics representation from interaction point clouds, robot states, and region-level proximity cues. The second stage freezes this representation and trains a Transformer-based diffusion policy, in which the predicted trajectory-stage probability adaptively modulates the proximity branch. This design allows the policy to emphasize global geometry during approach and exploit local proximity cues during dexterous manipulation. All training and inference are conducted on an Intel i9-14700K CPU and an NVIDIA RTX 4090D GPU.

\subsection{Latent Interaction Dynamics Architecture}
\label{app:latent_dynamics_architecture}

Given the interaction point cloud $o_t^p$ and robot state $s_t$, the observation encoder $E_\theta$ extracts geometric features and fuses them with the state embedding to obtain the observation latent $z_t^{obs}$. This latent is shared by the forward dynamics model (FDM), inverse differential module (IDM), and trajectory head in Stage 1, and is subsequently used as the global condition for policy generation in Stage 2.

Before learning the interaction dynamics, we independently pretrain a proximity variational autoencoder (VAE) for 100 epochs to compress the region-level proximity observation $o_t^{prox}$ into a compact latent representation. The encoder parameterizes a Gaussian posterior, while the decoder reconstructs the original proximity observation. In our implementation, the proximity input and latent dimensions are 124 and 16, respectively. The VAE is optimized using a reconstruction loss and a KL-divergence regularizer:
\begin{equation}
	\mathcal{L}_{\mathrm{vae}}
	=
	\lambda_{\mathrm{rec}}
	\left\|
	\hat{o}_t^{prox}-o_t^{prox}
	\right\|_2^2
	+
	\lambda_{\mathrm{KL}}
	D_{\mathrm{KL}}
	\left(
	q(z_t^{prox}\mid o_t^{prox})
	\,\|\,\mathcal{N}(0,I)
	\right),
\end{equation}
where $\lambda_{\mathrm{rec}}=1.0$ and $\lambda_{\mathrm{KL}}=0.01$. After pretraining, the VAE is frozen, and its posterior mean $\mu_{\mathrm{prox}}(o_t^{prox})$ is used as a deterministic proximity target for interaction dynamics learning.

The forward dynamics model $\mathrm{FDM}(\cdot)$ is implemented as a DiT-style conditional denoising model that predicts future observation features after action execution. Specifically, the future observation latent is perturbed and then denoised conditioned on the current latent and action sequence:
\begin{equation}
	\begin{aligned}
		\tilde{z}_{t+h}^{obs,(\ell)}
		&=
		\sqrt{\bar{\alpha}_\ell}z_{t+h}^{obs}
		+
		\sqrt{1-\bar{\alpha}_\ell}\epsilon, \\
		\hat{z}_{t+h}^{obs}
		&=
		f_\theta
		\left(
		\tilde{z}_{t+h}^{obs,(\ell)},\ell
		\mid z_t^{obs},a_{t:t+h-1}
		\right),
	\end{aligned}
\end{equation}
where $\ell$ denotes the diffusion step and $\epsilon$ is Gaussian noise. This forward-prediction objective encourages $z_t^{obs}$ to encode not only object geometry and hand configuration, but also short-horizon interaction dynamics associated with approach, grasping, and contact transfer.

To connect the predicted state transition with local geometric proximity, ProxiDex introduces the inverse differential module $\mathrm{IDM}(\cdot)$, implemented as an MLP with layer normalization and nonlinear activations. For one-step prediction, IDM maps the change in observation latent to the next-step proximity latent, which is aligned with the deterministic target provided by the frozen VAE:
\begin{equation}
	\begin{aligned}
		\hat{z}_{t+1}^{\mathrm{prox}}
		&=
		\mathrm{IDM}
		\left(
		\hat{z}_{t+1}^{\mathrm{obs}}
		-
		z_t^{\mathrm{obs}}
		\right), \\
		\mathcal{L}_{\mathrm{inv}}
		&=
		\left\|
		\hat{z}_{t+1}^{\mathrm{prox}}
		-
		\mu_{\mathrm{prox}}
		\left(
		o_{t+1}^{\mathrm{prox}}
		\right)
		\right\|_2^2.
	\end{aligned}
\end{equation}
This alignment encourages the observation-latent transition to preserve action-induced changes in hand--object proximity and enables IDM to infer implicit proximity information from visual and robot-state dynamics.

In parallel, the trajectory-stage head predicts the interaction probability $p_t^{\mathrm{int}}$ from $z_t^{obs}$ and is supervised by $o_t^{\mathrm{traj}}$, as defined in Appendix~\ref{app:geometric_proximity}. Together, forward prediction, proximity-latent alignment, and trajectory-stage classification produce a physically meaningful interaction representation for downstream control.

As summarized in Table~\ref{tab:dynamics_policy_hyperparameters}, the point-cloud feature and proximity latent dimensions are 128 and 16, respectively. The dynamics branch uses a hidden dimension of 256 and a short prediction horizon $H_{\mathrm{dyn}}=2$, allowing it to capture immediate action-induced interaction changes while limiting long-horizon prediction errors. Stage-1 training spans 301 epochs in total. We first pretrain the proximity VAE for 100 epochs. The VAE is then frozen, and the observation encoder, FDM, IDM, and trajectory head are trained for the remaining 201 epochs. The forward-dynamics, proximity-alignment, and trajectory-stage losses are assigned equal weights.

\begin{table}[!t]
	\centering
	\caption{Key training and model hyperparameters for the Interaction Dynamics Model.}
	\label{tab:dynamics_policy_hyperparameters}
	\small
	\renewcommand{\arraystretch}{1.15}
	\begin{tabular}{lcc}
		\toprule
		\textbf{Parameter} & \textbf{Symbol} & \textbf{Value} \\
		\midrule
		
		\multicolumn{3}{l}{\textit{Policy and Diffusion Settings}} \\
		Policy Horizon & $T$ & 16 \\
		Observation Horizon & $T_o$ & 4 \\
		Action Horizon & $T_a$ & 8 \\
		Diffusion Training Timesteps & $K$ & 100 \\
		
		\midrule
		\multicolumn{3}{l}{\textit{Dynamics and Proximity Modeling}} \\
		Point-Cloud Feature Dimension & $d_p$ & 128 \\
		Future Prediction Steps & $H_{\mathrm{dyn}}$ & 2 \\
		Proximity Latent Dimension & $d_{\mathrm{prox}}$ & 16 \\
		Action Embedding Dimension & $d_a$ & 16 \\
		FDM Hidden Dimension & $d_{\mathrm{fdm}}$ & 256 \\
		Proximity VAE Pretraining Epochs & -- & 100 \\
		Action Dropout / Negative Sampling Probability & -- & 0.1 / 0.1 \\
		Stage-1 Loss Weights
		& --
		& $\lambda_{\mathrm{v}}=\lambda_{\mathrm{f}}=\lambda_{\mathrm{c}}=\lambda_{\mathrm{t}}=1.0$ \\
		Stage-2 Loss Weights
		& --
		& $\lambda_{\mathrm{diff}}=1.0,\ \lambda_{\mathrm{gate}}=0.001$ \\
		Reconstruction / KL Weights
		& --
		& $\lambda_{\mathrm{rec}}=1.0,\ \lambda_{\mathrm{KL}}=0.01$ \\
		
		\midrule
		\multicolumn{3}{l}{\textit{Optimization Settings}} \\
		Optimizer & -- & AdamW \\
		Training Epochs & -- & 301 \\
		Batch Size & -- & 64 \\
		Learning Rate & -- & $5 \times 10^{-5}$ \\
		Weight Decay & -- & $1 \times 10^{-6}$ \\
		Learning Rate Schedule & -- & Cosine \\
		
		\bottomrule
	\end{tabular}
	\vspace{-10pt}
\end{table}

\subsection{Trajectory-Adaptive Proximity Policy Learning}
\label{app:trajectory_adaptive_policy}

The Stage-2 policy backbone is a Transformer-based diffusion model. The policy is trained using VAE-derived proximity latents, whereas IDM-estimated proximity latents are used during online deployment. It takes a
noised action sequence as input and denoises it under two types of conditions:
the observation latent $z_t^{obs}$, which provides global information such as
target geometry, object pose, and hand configuration; and the proximity latent
$z_t^{prox}$, which provides local distance cues between hand regions and the
object surface. To avoid overusing local feedback before contact, we gate the
proximity branch according to the interaction probability $p_t^{int}$ predicted
by the trajectory-stage head in Stage 1:
\begin{equation}
	\bar{z}_t^{prox}
	=
	g(p_t^{int})z_t^{prox},
	\qquad
	g(p)=1-\alpha+\alpha p,
	\qquad
	\alpha\in[0,1].
\end{equation}

During the first 2,000 training steps, we set $p_t^{int}=0.5$ to stabilize
policy optimization. After this warm-up period, the interaction probability is
computed as $p_t^{int}=\operatorname{sigmoid}(c_t/\tau)$, where $c_t$ is the
predicted interaction logit. The temperature $\tau$ is annealed from 2.0 to
0.7 over 50,000 steps, while $\alpha$ is linearly increased from 0 to 1 over
20,000 steps after warm-up. This schedule gradually introduces proximity
modulation without abruptly changing the policy condition.

When $p_t^{int}$ is low, the policy relies more on $z_t^{obs}$ for global
approach. As the interaction probability increases, the proximity branch is
strengthened, enabling the policy to refine fingertip closure, contact
maintenance, and release timing. To prevent the gate probability from
prematurely saturating, we introduce a negative binary-entropy regularizer:
\begin{equation}
	\begin{gathered}
		\mathcal{L}_{\mathrm{gate}}
		=
		\mathbb{E}\left[
		p_t^{\mathrm{int}}\log\left(p_t^{\mathrm{int}}+\epsilon\right)
		+
		\left(1-p_t^{\mathrm{int}}\right)
		\log\left(1-p_t^{\mathrm{int}}+\epsilon\right)
		\right], \\[2mm]
		\mathcal{L}_{\mathrm{stage2}}
		=
		\mathcal{L}_{\mathrm{diff}}
		+
		\lambda_{\mathrm{gate}}\mathcal{L}_{\mathrm{gate}}.
	\end{gathered}
\end{equation}
where $\mathcal{L}_{\mathrm{diff}}$ denotes the diffusion denoising loss and
$\epsilon$ is a small constant for numerical stability.

For temporal policy generation, the proximity condition is projected into the
policy embedding space and shared across the predicted action sequence, while
the observation latent remains available as stable global context. This
conditioning scheme preserves global planning ability while allowing the
interaction probability to adaptively regulate the contribution of local
proximity information.

The detailed hyperparameters for proximity-adaptive policy learning are
summarized in Table~\ref{tab:proximity_policy_hyperparameters}. Overall,
Stage 2 follows the same temporal window configuration as Stage 1 while using
fewer denoising steps during inference for efficient online deployment. Since
the Stage-1 representation is frozen, optimization mainly focuses on the
diffusion policy and its trajectory-adaptive proximity conditioning.

\begin{table}[!t]
	\centering
	\caption{Key training and model hyperparameters for the Proximity Policy.}
	\label{tab:proximity_policy_hyperparameters}
	\small
	\renewcommand{\arraystretch}{1.15}
	\begin{tabular}{lcc}
		\toprule
		\textbf{Parameter} & \textbf{Symbol} & \textbf{Value} \\
		\midrule
		
		\multicolumn{3}{l}{\textit{Policy and Diffusion Settings}} \\
		Policy Horizon
		& $T$
		& 16 \\
		Observation Horizon
		& $T_o$
		& 4 \\
		Action Horizon
		& $T_a$
		& 8 \\
		Diffusion Training Timesteps
		& $K$
		& 100 \\
		Inference Denoising Steps
		& --
		& 10 \\
		
		\midrule
		\multicolumn{3}{l}{\textit{Proximity Policy Modeling}} \\
		Point-Cloud Feature Dimension
		& $d_p$
		& 128 \\
		Proximity Latent Dimension
		& $d_{\mathrm{prox}}$
		& 16 \\
		Interaction Logit Dimension
		& $C_{\mathrm{int}}$
		& 1 \\
		Gate Warm-up Steps
		& $N_{\mathrm{warm}}$
		& 2,000 \\
		Gate Temperature Range
		& $\tau$
		& $2.0 \rightarrow 0.7$ \\
		Temperature Annealing Steps
		& $N_{\tau}$
		& 50,000 \\
		Gate Ramp-up Steps
		& $N_{\alpha}$
		& 20,000 \\
		Stage-2 Loss Weights
		& --
		& $\lambda_{\mathrm{diff}}=1.0,\ 
		\lambda_{\mathrm{gate}}=10^{-3}$ \\
		
		\midrule
		\multicolumn{3}{l}{\textit{Transformer Backbone}} \\
		Transformer Layers
		& $L$
		& 8 \\
		Attention Heads
		& $H$
		& 8 \\
		Embedding Dimension
		& $d_{\mathrm{emb}}$
		& 256 \\
		
		\midrule
		\multicolumn{3}{l}{\textit{Optimization Settings}} \\
		Optimizer
		& --
		& AdamW \\
		Training Epochs
		& --
		& 301 \\
		Batch Size
		& --
		& 128 \\
		Learning Rate
		& --
		& $1 \times 10^{-4}$ \\
		Weight Decay
		& --
		& $1 \times 10^{-6}$ \\
		
		\bottomrule
	\end{tabular}
	\vspace{-10pt}
\end{table}

\subsection{Dynamics-Consistency Guided Policy Inference}
\label{app:closed_loop_inference}

During online inference, ProxiDex places policy generation and FDM prediction in the same loop. The system encodes the real-time interaction point cloud $o_t^p$ and robot state $s_t$ into $z_t^{obs}$, estimates $\hat{z}_t^{\mathrm{prox}}$ through IDM, and obtains $p_t^{\mathrm{int}}$ from the trajectory head. The diffusion policy then generates an action sequence and executes actions in a receding-horizon manner.

After action execution, FDM predicts the next latent state and compares it with the encoded real observation:
\begin{equation}
	e_{t+1}^{\mathrm{dyn}}=\left\|z_{t+1}^{obs}-\mathrm{FDM}(z_t^{obs},a_t)\right\|_2 .
\end{equation}
If $e_{t+1}^{\mathrm{dyn}}\le \tau_{\rm dyn}$, the system continues closed-loop control with real-time perception. If the deviation exceeds the threshold, the system treats the current visual feedback as potentially affected by occlusion, pose drift, or short-term tracking failure, and temporarily switches to dynamics takeover, using autoregressive FDM latents to maintain action generation. Once the deviation returns below the threshold, the system switches back to real observations and re-aligns the state.

This mechanism makes FDM both a training-time dynamics supervisor and an inference-time physical consistency checker. In deployment, FoundationPose++ provides object-pose updates at around $10$ Hz, the Proximity Policy runs at about $28$ Hz, and the learned world model runs at about $20$ Hz. These runtime rates are sufficient for real-time closed-loop inference, enabling dynamics takeover to maintain action generation from predicted latents when visual pose tracking becomes temporarily unreliable. Algorithm~\ref{alg:closed_loop_inference} summarizes the procedure.

\begin{algorithm}[!t]
	\caption{Dynamics-Consistency Guided Policy Inference}
	\label{alg:closed_loop_inference}
	\scriptsize
	\begin{algorithmic}[1]
		\Require Multimodal observations $\{(P_t,S_t)\}_{t=0}^{T+1}$,
		encoder $E$, forward dynamics model $\mathrm{FDM}$,
		diffusion policy $\pi_\theta$, dynamics threshold $\tau_{\mathrm{dyn}}$,
		and scaling factor $\alpha$
		\Ensure Executed action sequence $\{a_t\}_{t=1}^{T}$
		
		\State $z_0^{\mathrm{obs}} \gets E(P_0,S_0)$
		\State $z_0^{\mathrm{in}} \gets z_0^{\mathrm{obs}}$
		\State $\mathrm{Mode} \gets \mathrm{Normal}$
		
		\For{$t=1$ \textbf{to} $T$}
		\Statex \textit{/* Observation encoding */}
		\State $z_t^{\mathrm{obs}} \gets E(P_t,S_t)$
		\If{$\mathrm{Mode}=\mathrm{Normal}$}
		\State $z_t^{\mathrm{in}} \gets z_t^{\mathrm{obs}}$
		\Else
		\State $z_t^{\mathrm{in}} \gets \hat{z}_t^{\mathrm{obs}}$
		\EndIf
		\State $\Delta z_t
		\gets z_t^{\mathrm{in}}-z_{t-1}^{\mathrm{in}}$
		
		\Statex \textit{/* Current proximity inference and adaptive gating */}
		\State $\hat{z}_t^{\mathrm{prox}}
		\gets \mathrm{IDM}(\Delta z_t)$
		\State $p_t^{\mathrm{int}}
		\gets h(z_t^{\mathrm{in}})$
		\State $\bar{z}_t^{\mathrm{prox}}
		\gets
		\left(1-\alpha+\alpha p_t^{\mathrm{int}}\right)
		\hat{z}_t^{\mathrm{prox}}$
		
		\Statex \textit{/* Action generation and latent dynamics prediction */}
		\State $a_t
		\gets
		\pi_\theta
		\left(
		z_t^{\mathrm{in}},
		\bar{z}_t^{\mathrm{prox}}
		\right)$
		\State Execute $a_t$ via the robot controller
		\State $\hat{z}_{t+1}^{\mathrm{obs}}
		\gets
		\mathrm{FDM}
		\left(
		z_t^{\mathrm{in}},a_t
		\right)$
		
		\Statex \textit{/* Dynamics monitoring and mode switching */}
		\State Receive the next observation $(P_{t+1},S_{t+1})$
		\State $z_{t+1}^{\mathrm{obs}}
		\gets E(P_{t+1},S_{t+1})$
		\State $e_{t+1}^{\mathrm{dyn}}
		\gets
		\left\|
		z_{t+1}^{\mathrm{obs}}
		-
		\hat{z}_{t+1}^{\mathrm{obs}}
		\right\|_2$
		
		\If{$e_{t+1}^{\mathrm{dyn}}\leq\tau_{\mathrm{dyn}}$}
		\State $\mathrm{Mode}\gets\mathrm{Normal}$
		\State $z_{t+1}^{\mathrm{in}}
		\gets z_{t+1}^{\mathrm{obs}}$
		\Else
		\State $\mathrm{Mode}\gets\mathrm{Dream}$
		\State $z_{t+1}^{\mathrm{in}}
		\gets\hat{z}_{t+1}^{\mathrm{obs}}$
		\EndIf
		\EndFor
	\end{algorithmic}
\end{algorithm}

\FloatBarrier

\section{Simulation Experiments}
\label{app:simulation_setup}

\subsection{Simulation Setup and Task Description}
\label{app:simulation_tasks}

We evaluate ProxiDex on Adroit, DexArt, and three self-designed IsaacLab tasks, as illustrated in Fig.~\ref{fig:simulation_tasks}. Adroit covers classic dexterous manipulation tasks requiring fine finger coordination, such as Hammer, Door, and Pen, while DexArt provides articulated-object interactions closer to daily manipulation, including Laptop, Faucet, Toilet, and Bucket. Since the simulator directly provides object poses and robot joint states, we can reliably synthesize occlusion-free interaction point clouds $o_t^p$ and compute region-level proximity observations $o_t^{\mathrm{prox}}$. The interaction point clouds are visualized in Fig.~\ref{fig:simulation_point_clouds}, showing that they preserve active hand--object regions throughout manipulation.

The self-designed tasks are defined as follows:
\begin{itemize}
	\setlength{\itemsep}{2pt}
	\setlength{\parsep}{0pt}
	\item \textbf{Box.} The robot grasps a box-like object with multiple fingers and places it into a basket. This task evaluates grasp stability and collision-free transport around the basket rim.
	\item \textbf{Driller.} The robot side-grasps a drill and moves the drill tip above a target hole. This task requires stable tool orientation and accurate alignment after long-range motion.
	\item \textbf{Banana.} The robot pinches a banana-like object using the thumb and index finger and places it into a bowl. This task is sensitive to fingertip closure timing, contact location, and holding stability.
\end{itemize}

Together, these tasks cover rigid-object grasping, articulated-object manipulation, tool alignment, and fine fingertip pinching. The implementation details below keep the data split and evaluation protocol consistent across all methods, so that the comparisons isolate the effect of interaction point clouds, proximity features, and dynamics modeling.

\subsection{Simulation Implementation Details}
\label{app:simulation_details}

\begin{figure}[!t]
	\centering
	\captionsetup{skip=4pt}
	\includegraphics[width=0.9\textwidth]{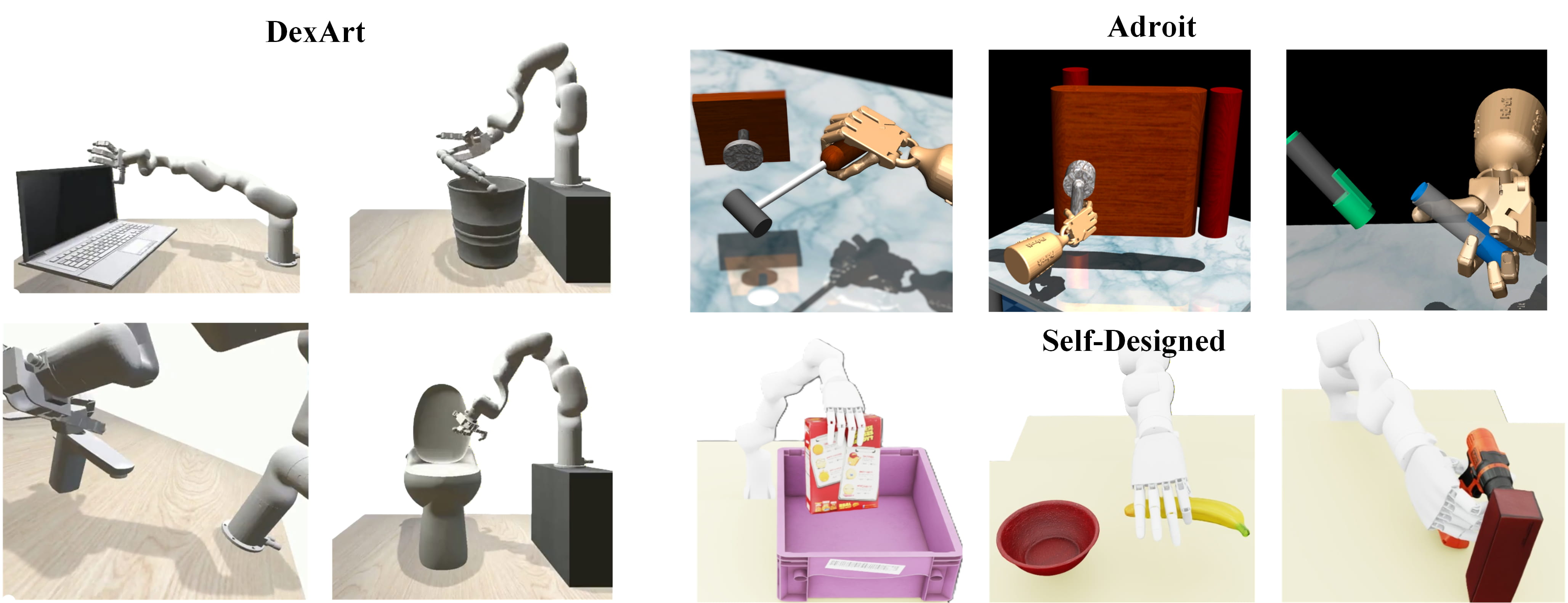}
	\caption{Visualization of the simulation tasks. 
		We evaluate ProxiDex on ten dexterous manipulation tasks, including four DexArt tasks, three Adroit tasks, and three custom tasks built in IsaacLab.}
	\label{fig:simulation_tasks}
	
\end{figure}

For Adroit and DexArt, demonstrations are generated by reinforcement-learning experts, while the self-designed IsaacLab demonstrations are collected through VR teleoperation. All methods use the same demonstration split, action horizon, and evaluation protocol. We report the mean success rate and standard deviation across repeated evaluations. Fig.~\ref{fig:simulation_tasks} visualizes the ten simulation tasks before the quantitative comparison, and Table~\ref{tab:algorithm_detail_performance} provides the corresponding task-level results.

\begin{table*}[!t]
	\centering
	\caption{Simulation results on dexterous manipulation benchmarks. We evaluate all methods on Adroit, DexArt, and three self-designed IsaacLab tasks.}
	\label{tab:algorithm_detail_performance}
	\scriptsize
	\renewcommand{\arraystretch}{1.15}
	\setlength{\tabcolsep}{2pt}
	\resizebox{\textwidth}{!}{
		\begin{tabular}{c|ccc|cccc|ccc|c}
			\specialrule{1.2pt}{0pt}{0pt}
			\multirow{2}{*}{Algorithm\textbackslash{}Tasks} 
			& \multicolumn{3}{c|}{Adroit} 
			& \multicolumn{4}{c|}{DexArt} 
			& \multicolumn{3}{c|}{Self-Designed} 
			& \multirow{2}{*}{Avg.} \\ 
			\cline{2-11}
			& Hammer & Door & Pen 
			& Laptop & Faucet & Toilet & Bucket 
			& Box & Driller & Banana 
			& \\ 
			\hline
			
			DP3 
			& \cellcolor{gray!20}\textbf{100.0$\pm$0.0} 
			& 76.7$\pm$4.7 
			& 56.7$\pm$2.6 
			& 89.7$\pm$0.9 
			& 41.7$\pm$0.5 
			& 79.7$\pm$0.9 
			& 31.3$\pm$0.5 
			& 90.3$\pm$3.2 
			& 74.3$\pm$2.6 
			& 77.8$\pm$1.2 
			& 71.8$\pm$1.7 \\ 
			
			ManiFlow 
			& \cellcolor{gray!20}\textbf{100.0$\pm$0.0} 
			& 80.3$\pm$1.2 
			& 55.5$\pm$5.8 
			& \cellcolor{gray!20}\textbf{93.0$\pm$1.6} 
			& 45.0$\pm$3.6 
			& 79.9$\pm$3.3 
			& 35.3$\pm$2.1 
			& 94.7$\pm$2.1 
			& 78.6$\pm$1.4 
			& 79.8$\pm$2.3 
			& 74.2$\pm$2.3 \\ 
			
			AFRO 
			& 96.0$\pm$2.8 
			& 83.8$\pm$3.4 
			& 72.3$\pm$2.1 
			& 86.3$\pm$2.5 
			& 42.3$\pm$1.8 
			& 82.2$\pm$3.4 
			& 48.0$\pm$2.3 
			& \cellcolor{gray!20}\textbf{97.2$\pm$1.3} 
			& 79.4$\pm$1.5 
			& 81.4$\pm$1.8 
			& 76.9$\pm$2.3 \\ 
			
			CordViP 
			& 96.0$\pm$2.1 
			& 84.7$\pm$1.7 
			& 76.2$\pm$3.2 
			& 92.6$\pm$2.1 
			& \cellcolor{gray!20}\textbf{56.5$\pm$1.3} 
			& 81.0$\pm$2.2 
			& 57.0$\pm$1.6 
			& 96.3$\pm$2.0 
			& 85.3$\pm$1.1 
			& 84.7$\pm$1.6 
			& 81.0$\pm$1.9 \\ 
			
			\textbf{ProxiDex (Ours)} 
			& \cellcolor{gray!20}\textbf{100.0$\pm$0.0} 
			& \cellcolor{gray!20}\textbf{86.5$\pm$1.9} 
			& \cellcolor{gray!20}\textbf{84.3$\pm$2.4} 
			& \cellcolor{gray!20}\textbf{93.0$\pm$1.2} 
			& 53.4$\pm$2.1 
			& \cellcolor{gray!20}\textbf{84.3$\pm$2.3} 
			& \cellcolor{gray!20}\textbf{64.0$\pm$3.2} 
			& 96.7$\pm$1.6 
			& \cellcolor{gray!20}\textbf{89.2$\pm$2.0} 
			& \cellcolor{gray!20}\textbf{87.3$\pm$1.5} 
			& \cellcolor{gray!20}\textbf{83.9$\pm$1.8} \\ 
			
			\specialrule{1.2pt}{0pt}{0pt}
		\end{tabular}
	}
	\vspace{-10pt}
\end{table*}

The detailed task-level results in Table~\ref{tab:algorithm_detail_performance} further support the overall advantage of ProxiDex. ProxiDex achieves the best ten-task average of $83.9\%$, outperforming DP3, ManiFlow, AFRO, and CordViP by $12.1$, $9.7$, $7.0$, and $2.9$ percentage points, respectively. The improvement is consistent across Adroit, DexArt, and the self-designed tasks, indicating that proximity-aware dynamics benefit both standard dexterous benchmarks and customized tabletop manipulation scenarios.

A closer look at the task results shows that the gains are more evident in contact-rich tasks requiring sustained fingertip adjustment, such as Pen, Bucket, Driller, and Banana. These tasks involve continuous hand-object proximity regulation after initial contact, where the proposed proximity representation and adaptive policy modulation provide more informative feedback than geometry-only baselines. Although CordViP performs slightly better on Faucet, ProxiDex achieves the best aggregate performance by improving tasks with more complex contact-state transitions.

\begin{figure}[!t]
	\centering
	\captionsetup{skip=4pt}
	\includegraphics[width=0.9\textwidth]{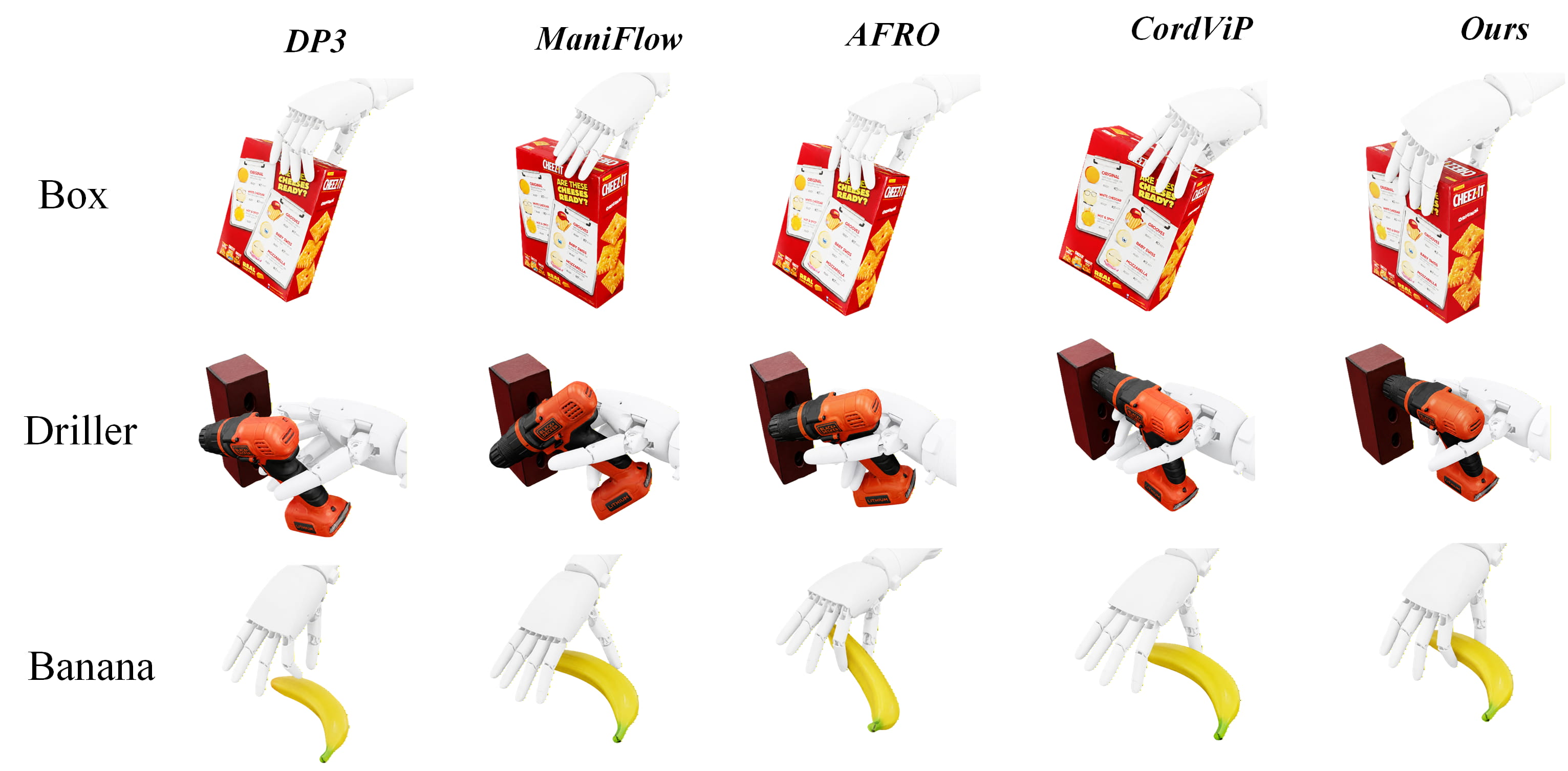}
	\caption{Visualization of hand--object interactions on the self-designed tasks, compared with typical failure cases from existing baselines.}
	\label{fig:simulation_qualitative}
	\vspace{-10pt}
\end{figure}

Fig.~\ref{fig:simulation_qualitative} provides qualitative evidence complementary to the quantitative results in Table~\ref{tab:algorithm_detail_performance}, comparing representative executions on the self-designed tasks with typical failure cases of the baselines. Specifically, in the Box task, baseline methods often rely on a small distal contact region and can lose grasp stability during transport, whereas ProxiDex adjusts finger closure after approaching the target. In Driller, visible-point-cloud methods struggle to maintain the relative relation between the drill and the hole under occlusion; ProxiDex improves local adjustment near the hole through proximity-difference constraints. In Banana, ProxiDex forms more suitable fingertip contact regions, showing the value of region-level proximity for fine pinching.

\FloatBarrier

\section{Additional Experiments and Analysis}
\label{app:additional_analysis}

\begin{figure}[!t]
	\centering
	\captionsetup{skip=4pt}
	\includegraphics[width=0.8\textwidth]{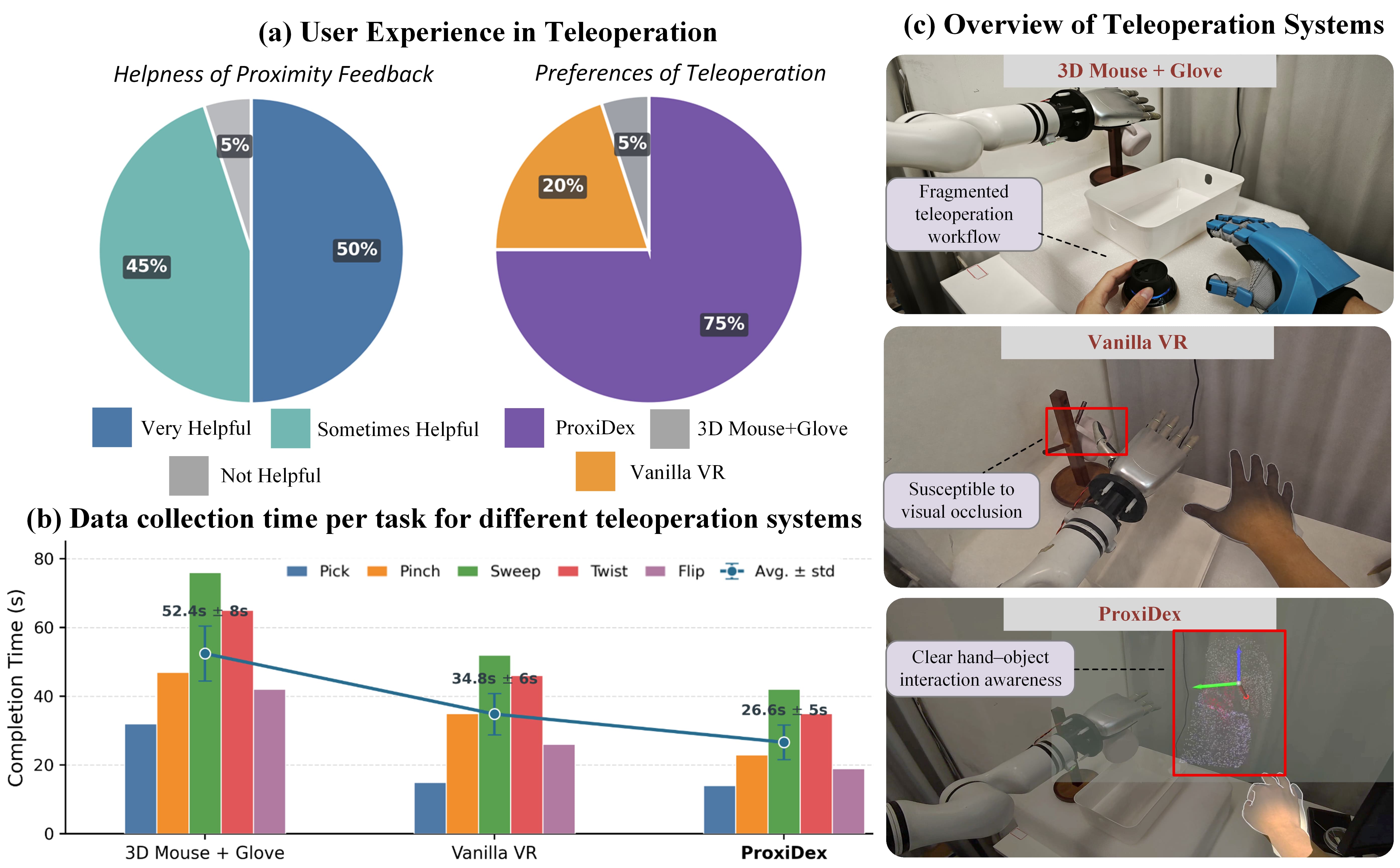}
	\caption{User study and teleoperation-system comparison. (a) User experience and system preference. (b) Task completion time. (c) Overview of the three teleoperation systems.}
	\label{fig13}
	\vspace{-10pt}
\end{figure}

\subsection{Teleoperation System Comparison}
\label{Teleoperation System Comparison}
To evaluate the benefit of immersive proximity feedback, we invited 20 participants to experience the three teleoperation systems shown in Fig.~\ref{fig13}(c). After 10 minutes of training, participants collected 30 demonstrations under each setting. As shown in Fig.~\ref{fig13}(a)(b), most participants found proximity feedback helpful, and 75\% preferred ProxiDex. ProxiDex also reduces the average task completion time to $26.6\pm5$ s, outperforming the other two systems. This advantage mainly comes from more direct interaction feedback: compared with Vanilla VR, which is susceptible to hand--object occlusion, ProxiDex visualizes hand--object distances and spatial relations through color-coded proximity point clouds, enabling timely adjustment of grasp position and finger closure and thereby improving teleoperation experience and demonstration-collection efficiency.

\subsection{Ablation Study}
\label{app:ablation_study}

\textbf{Ablation on core components.} Table~\ref{tab:component_ablation} reports the task-level numerical results corresponding to the ablation summary in Fig.~\ref{fig4_55}. The largest average drops occur when geometric interaction inputs are removed: removing interaction point clouds decreases the success rate from $77.6\%$ to $70.1\%$, while removing proximity cues decreases it to $70.6\%$, with drops of $7.5$ and $7.0$ points, respectively. The degradation is especially clear on contact-sensitive tasks: Bucket drops by $15.1$ and $13.4$ points under these two removals, while Pen drops by $8.4$ points without interaction point clouds. These results indicate that complete hand--object geometry and region-level proximity provide complementary cues for contact-rich manipulation.

\begin{table*}[]
	\centering
	\caption{Ablation study of key components on representative dexterous manipulation tasks.}
	\label{tab:component_ablation}
	\scriptsize
	\renewcommand{\arraystretch}{1.15}
	\setlength{\tabcolsep}{3pt}
	\resizebox{\textwidth}{!}{
		\begin{tabular}{c|cccccc|c}
			\specialrule{1.2pt}{0pt}{0pt}
			Components & Door & Pen & Laptop & Faucet & Toilet & Bucket & Avg. \\ 
			\hline
			
			w/o Prox 
			& 76.7$\pm$2.3 
			& 79.4$\pm$1.6 
			& 88.3$\pm$1.5 
			& 47.9$\pm$1.5 
			& 80.7$\pm$1.6 
			& 50.6$\pm$1.3 
			& 70.6$\pm$1.6 \\ 
			
			w/o Int. PC 
			& 80.3$\pm$1.2 
			& 75.9$\pm$2.4 
			& 86.1$\pm$1.4 
			& 48.7$\pm$1.4 
			& 80.4$\pm$1.5 
			& 48.9$\pm$2.0 
			& 70.1$\pm$1.7 \\ 
			
			w/o FDM 
			& 83.8$\pm$2.2 
			& 80.3$\pm$1.7 
			& 89.6$\pm$1.5 
			& 50.5$\pm$1.7 
			& 81.1$\pm$1.7 
			& 56.4$\pm$1.8 
			& 73.6$\pm$1.8 \\ 
			
			w/o IDM 
			& 84.7$\pm$1.7 
			& 80.9$\pm$2.1 
			& 92.3$\pm$1.7 
			& 51.8$\pm$1.3 
			& 82.0$\pm$1.8 
			& 60.1$\pm$1.5 
			& 75.3$\pm$1.7 \\ 
			
			w/o Closed-Loop 
			& 85.2$\pm$1.2 
			& 82.5$\pm$1.2 
			& 92.8$\pm$1.3 
			& 52.6$\pm$1.1 
			& 83.4$\pm$2.0 
			& 62.3$\pm$1.6 
			& 76.5$\pm$1.4 \\ 
			
			\textbf{ProxiDex (Ours)} 
			& \cellcolor{gray!20}\textbf{86.5$\pm$1.9} 
			& \cellcolor{gray!20}\textbf{84.3$\pm$2.4} 
			& \cellcolor{gray!20}\textbf{93.0$\pm$1.2} 
			& \cellcolor{gray!20}\textbf{53.4$\pm$2.1} 
			& \cellcolor{gray!20}\textbf{84.3$\pm$2.3} 
			& \cellcolor{gray!20}\textbf{64.0$\pm$3.2} 
			& \cellcolor{gray!20}\textbf{77.6$\pm$2.2} \\ 
			
			\specialrule{1.2pt}{0pt}{0pt}
		\end{tabular}
	}
	\vspace{-10pt}
\end{table*}

From the dynamics perspective, removing FDM reduces the average success rate to $73.6\%$, corresponding to a $4.0$-point drop, while removing IDM reduces it to $75.3\%$, a $2.3$-point drop. FDM contributes more strongly to tasks that require predicting near-future contact consequences, whereas IDM improves sensitivity to local proximity changes caused by actions. Removing closed-loop feedback yields $76.5\%$, a smaller but consistent $1.1$-point drop. This pattern suggests that most of the performance gain comes from representation and proximity-aware policy learning, while inference-time dynamics checking mainly stabilizes execution when perception is temporarily unreliable.

The component-level ablation motivates a separate analysis of the Stage-2 policy architecture: after the interaction representation is fixed, the remaining question is whether the policy backbone can use proximity information at the correct manipulation phase.

\begin{figure}[!t]
	\centering
	\captionsetup{skip=4pt}
	\includegraphics[width=1.0\textwidth]{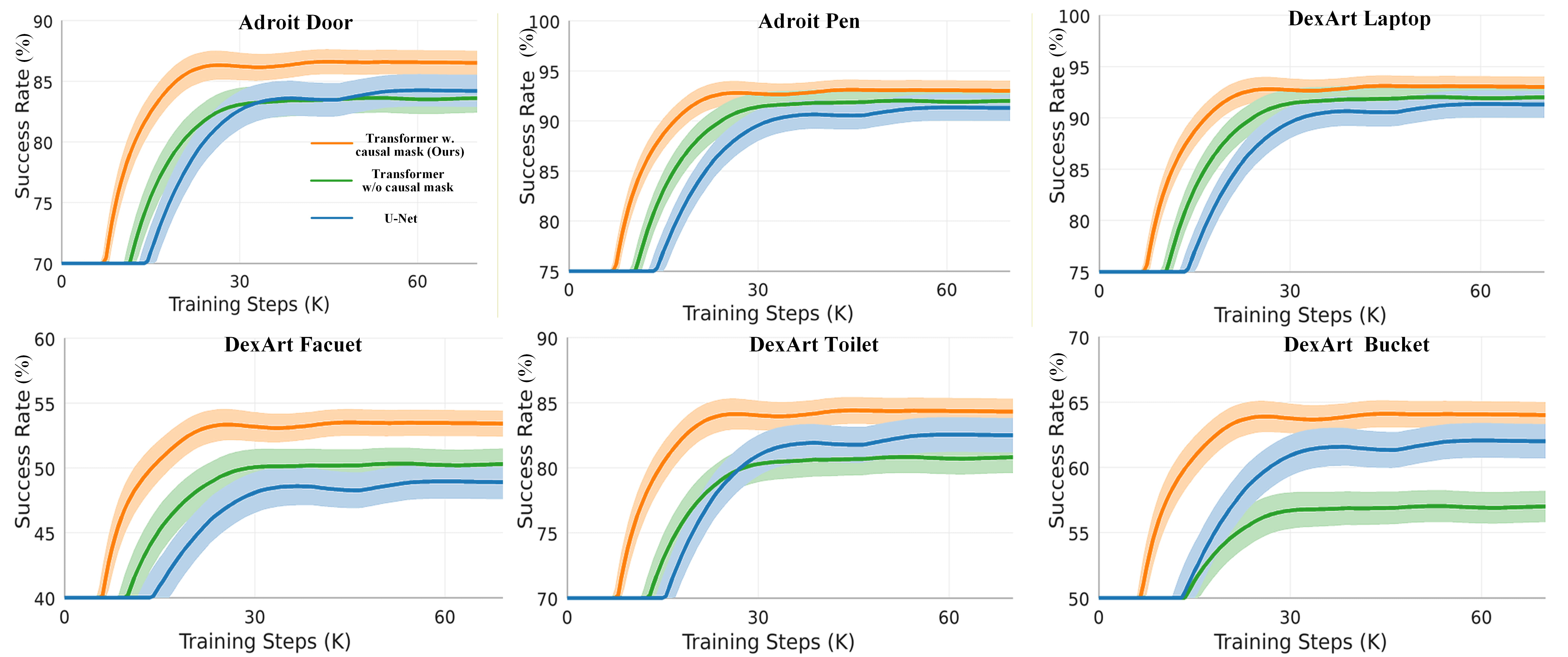}
	\caption{Policy-backbone ablation study on six Adroit and DexArt tasks. 
		The proposed causal Transformer with adaptive weight converges faster and achieves higher final success rates.}
	\label{fig:policy_backbone_ablation}
	\vspace{-10pt}
\end{figure}

\textbf{Ablation on the policy backbone.} Complementing the real-robot backbone ablation in Table~\ref{tab1:task_success}, Fig.~\ref{fig:policy_backbone_ablation} further evaluates the Stage-2 policy backbone in simulation while keeping Stage-1 pretraining unchanged. Replacing the proposed causal Transformer with a U-Net reduces the average success rate by about $3.1$ percentage points, since temporal convolutions are less flexible in fusing multimodal tokens across manipulation phases. Replacing it with a standard Transformer reduces performance by about $2.8$ percentage points, showing that causal proximity masking and trajectory-adaptive weighting are important for stable contact-aware action generation. These drops are smaller than the $7.0$--$7.5$ point drops caused by removing proximity cues or interaction point clouds in Table~\ref{tab:component_ablation}, suggesting that the representation provides the primary contact information, while the backbone determines how effectively it is used over time.

\textbf{Ablation on the latent dynamics prediction horizon.} To examine how the dynamics prediction horizon affects latent dynamics modeling, we vary $H_{\mathrm{dyn}}$ and report the results in Table~\ref{tab:hdyn_ablation}. $H_{\mathrm{dyn}}=2$ achieves the best average success rate of $77.6\%$, slightly outperforming the other settings. Although the differences are modest, shorter horizons provide limited supervision for contact evolution, while overly long horizons may accumulate prediction errors and shift the dynamics objective toward coarse trajectory fitting. We therefore use $H_{\mathrm{dyn}}=2$ by default, as it balances local contact sensitivity and prediction stability.

\begin{table*}[!t]
	\centering
	\caption{Ablation study on the latent dynamics prediction horizon $H_{\mathrm{dyn}}$.}
	\label{tab:hdyn_ablation}
	\scriptsize
	\renewcommand{\arraystretch}{1.15}
	\setlength{\tabcolsep}{2pt}
	\resizebox{0.92\textwidth}{!}{
		\begin{tabular}{c|cccccc|c}
			\specialrule{1.2pt}{0pt}{0pt}
			$H_{\mathrm{dyn}}$ & Door & Pen & Laptop & Faucet & Toilet & Bucket & Avg. \\ 
			\hline
			
			1 
			& \cellcolor{gray!20}\textbf{86.8$\pm$1.7} 
			& 84.0$\pm$2.2 
			& 92.6$\pm$1.5 
			& 53.3$\pm$1.9 
			& 80.7$\pm$1.6 
			& 64.1$\pm$1.8 
			& 76.9$\pm$1.8 \\ 
			
			2 
			& 86.5$\pm$1.9 
			& \cellcolor{gray!20}\textbf{84.3$\pm$2.4} 
			& \cellcolor{gray!20}\textbf{93.0$\pm$1.2} 
			& 53.4$\pm$2.1 
			& \cellcolor{gray!20}\textbf{84.3$\pm$2.3} 
			& 64.0$\pm$3.2 
			& \cellcolor{gray!20}\textbf{77.6$\pm$2.2} \\ 
			
			4 
			& 85.9$\pm$2.0 
			& 83.6$\pm$1.9 
			& \cellcolor{gray!20}\textbf{93.0$\pm$1.5} 
			& \cellcolor{gray!20}\textbf{53.7$\pm$1.4} 
			& 81.1$\pm$1.7 
			& \cellcolor{gray!20}\textbf{64.3$\pm$1.9} 
			& 76.9$\pm$1.7 \\ 
			
			8 
			& 84.7$\pm$1.1 
			& 82.8$\pm$1.8 
			& 92.1$\pm$1.4 
			& 53.4$\pm$1.5 
			& 82.0$\pm$1.8 
			& 63.9$\pm$1.7 
			& 76.5$\pm$1.6 \\ 
			
			10 
			& 84.8$\pm$1.2 
			& 82.6$\pm$1.3 
			& 91.8$\pm$1.1 
			& 52.8$\pm$1.3 
			& 83.4$\pm$2.0 
			& 63.2$\pm$2.0 
			& 76.4$\pm$1.5 \\ 
			
			\specialrule{1.2pt}{0pt}{0pt}
		\end{tabular}
	}
	\vspace{-10pt}
\end{table*}

\FloatBarrier

\subsection{Data Scalability Evaluation}
\label{app:data_scalability}

The number of demonstrations directly affects how well an imitation-learning policy covers contact patterns. To evaluate learning efficiency under few-shot and medium-scale data regimes, we vary the number of expert trajectories on six Adroit and DexArt tasks and compare DP3, ManiFlow, AFRO, CordViP, and ProxiDex. The demonstration sizes are $N\in\{5,10,20,30,50,100\}$, and all other training settings remain unchanged.

\begin{figure}[!t]
	\centering
	\captionsetup{skip=4pt}
	\includegraphics[width=1.0\textwidth]{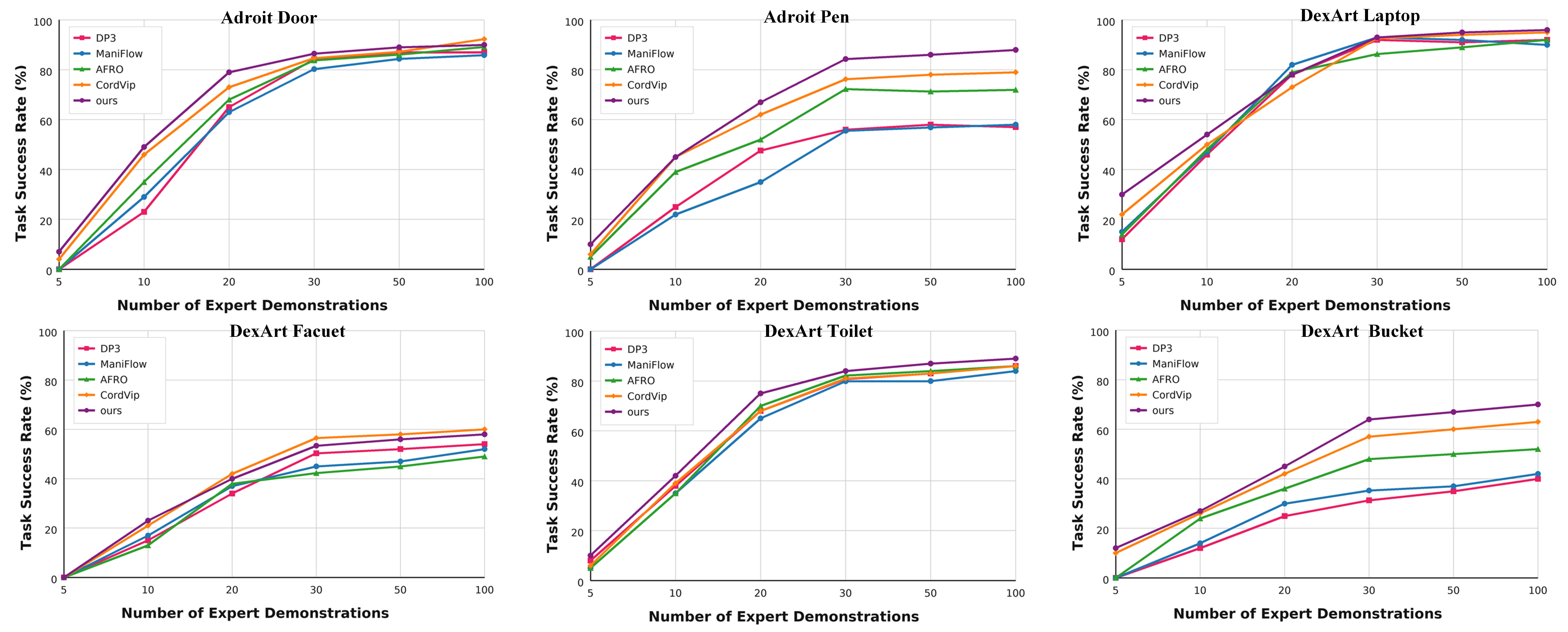}
	\caption{Scalability analysis with varying numbers of expert demonstrations on Adroit and DexArt tasks. 
		ProxiDex consistently outperforms the baselines from 5 to 100 demonstrations, demonstrating scalable and data-efficient learning from collected demonstrations.}
	\label{fig:data_scalability}
	\vspace{-10pt}
\end{figure}

As shown in Fig.~\ref{fig:data_scalability}, overall success rates improve as the number of demonstrations increases, but different methods exhibit different growth rates and saturation levels. DP3 and ManiFlow are unstable with limited data, suggesting that policies based only on visible scene point clouds require more samples to cover contact establishment and object-pose variation. AFRO improves sample efficiency through dynamics pretraining, but because it lacks explicit proximity supervision, its gains remain limited on strongly contact-driven tasks such as Bucket and Faucet. CordViP improves low-data performance through interaction point clouds, showing that recovering occlusion-free hand-object geometry is critical for few-shot learning.

ProxiDex achieves the highest average success rate across data scales. With $5$, $10$, $20$, $30$, $50$, and $100$ demonstrations, its six-task average success rates are approximately $11.5\%$, $40.0\%$, $64.0\%$, $77.5\%$, $80.0\%$, and $81.8\%$, respectively. The largest marginal gains occur before $30$ demonstrations: the average increases by about $28.5$ points from $5$ to $10$ demonstrations, $24.0$ points from $10$ to $20$, and $13.5$ points from $20$ to $30$. After $30$ demonstrations, the curve begins to saturate, gaining only $4.3$ points when moving from $30$ to $100$ demonstrations. Thus, $30$ demonstrations already recover about $94.7\%$ of the $100$-demonstration performance, indicating that proximity representation provides a strong structured inductive bias for learning finger closure, contact maintenance, and release timing from limited data.

\FloatBarrier

\subsection{Attention Visualization in Simulation}
\label{app:attention_visualization}

\begin{figure}[!t]
	\centering
	\captionsetup{skip=4pt}
	\includegraphics[width=1.0\textwidth]{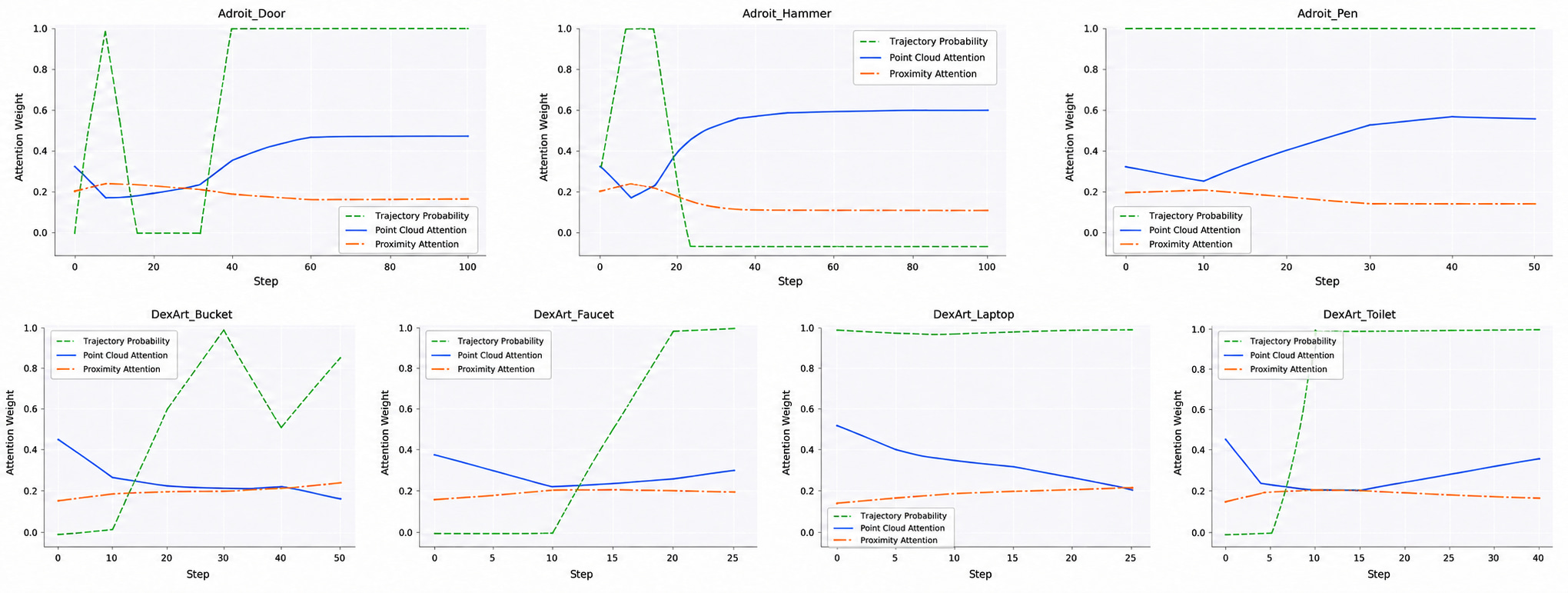}
	\caption{Attention weights visualization in simulation. We visualize the trajectory probability and averaged Transformer attention weights of point-cloud and proximity tokens across representative Adroit and DexArt tasks.}
	\label{fig:simulation_attention}
	\vspace{-10pt}
\end{figure}

We further visualize the attention weights in simulation environments, as shown in Fig.~\ref{fig:simulation_attention}. The trajectory probability usually rises rapidly before contact or during the early contact-establishment stage, indicating that the policy can identify the transition from approach to manipulation. Point-cloud attention mainly reflects external geometric observations and contributes more to estimating the target pose, hand--object configuration, and reachability, so it is more active during early motion and pose adjustment. In contrast, proximity attention remains smoother and acts as a near-field interaction modulation signal, constraining hand--object distance, contact stability, and manipulation robustness after contact becomes likely.

This visualization is consistent with the quantitative ablations above: the point-cloud branch provides global scene context, while the proximity branch becomes most useful when the task requires local contact refinement. Along with the change of trajectory probability, the policy adaptively balances attention between point-cloud geometry and proximity cues, leading to more stable dexterous control during contact establishment and subsequent manipulation.

\begin{wrapfigure}{r}{0.45\textwidth}
	\vspace{-8pt}
	\centering
	\includegraphics[width=0.4\textwidth]{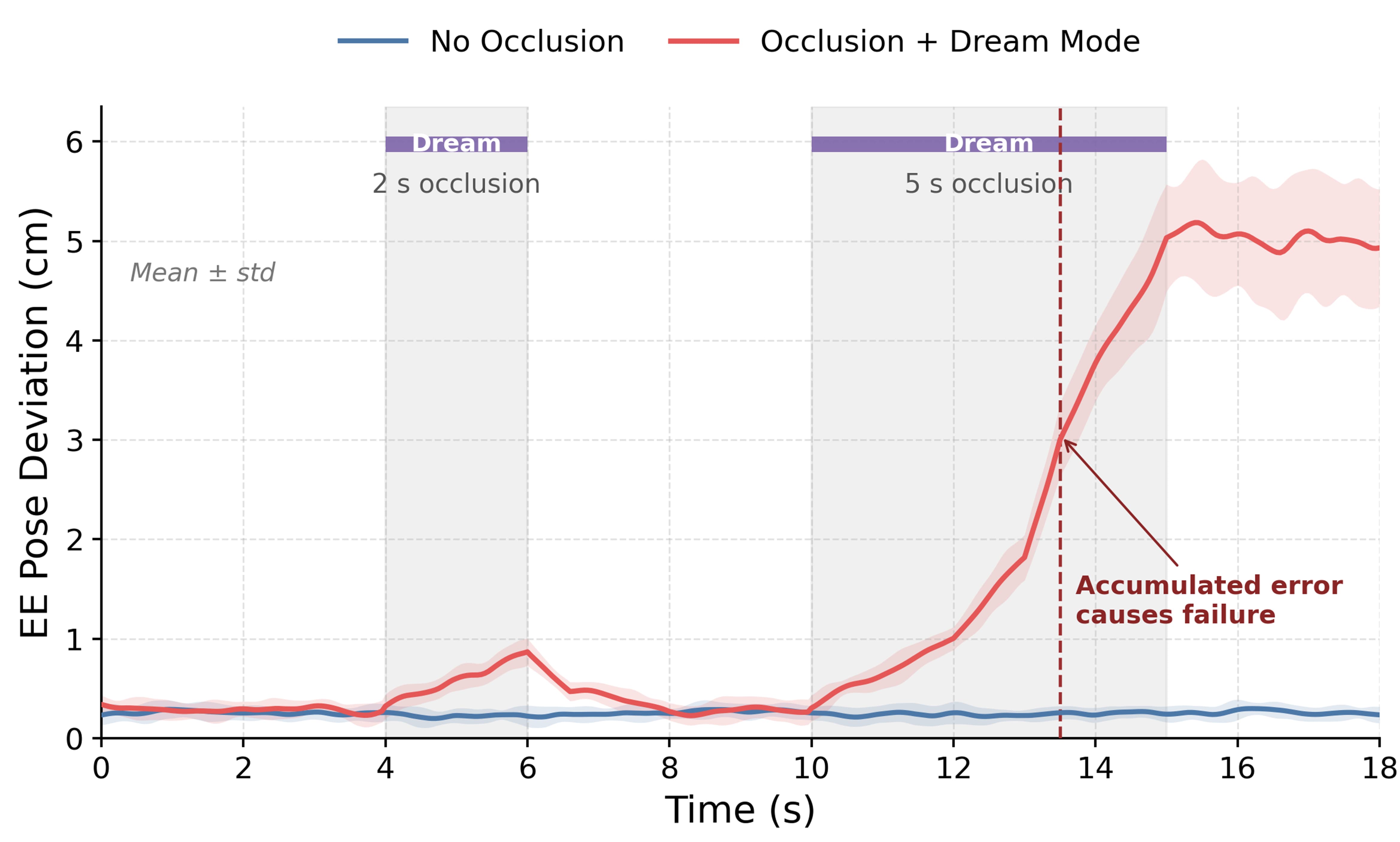}
	\caption{Dream-Mode robustness under controlled visual occlusion.}
	\label{fig17}
	\vspace{-10pt}
\end{wrapfigure}

\subsection{Dream-Mode Robustness Analysis}
\label{Dream-Mode Robustness Analysis}

To further characterize the stability range and error accumulation of Dream Mode under visual interruption, we measure end-effector pose deviations from the no-occlusion trajectory under controlled camera occlusions. As shown in Fig.~\ref{fig17}, a 2-s occlusion causes only a transient deviation of about 1 cm, after which the system gradually realigns once visual tracking recovers. In contrast, under a 5-s occlusion, prediction error accumulates continuously: the deviation grows slowly at first, but increases rapidly after about $3.5$ s to around 3 cm and further reaches about 5 cm, eventually causing arm drift and stalling. This indicates a reliable open-loop prediction horizon of approximately 3.5 s, which is sufficient for brief perception interruptions but not prolonged visual loss. This trend is also consistent with Fig.~\ref{fig5}, where recovery performance decreases as occlusion duration increases. During inference, FDM evaluates latent-dynamics consistency at 20 Hz, and $\tau_{\mathrm{dyn}}$ is set to the observation-feature discrepancy that first corresponds to a 1-cm end-effector pose deviation, enabling switching between normal visual feedback and dynamics takeover.

\begin{wrapfigure}{r}{0.43\textwidth}
	\vspace{-8pt}
	\centering
	\includegraphics[width=0.4\textwidth]{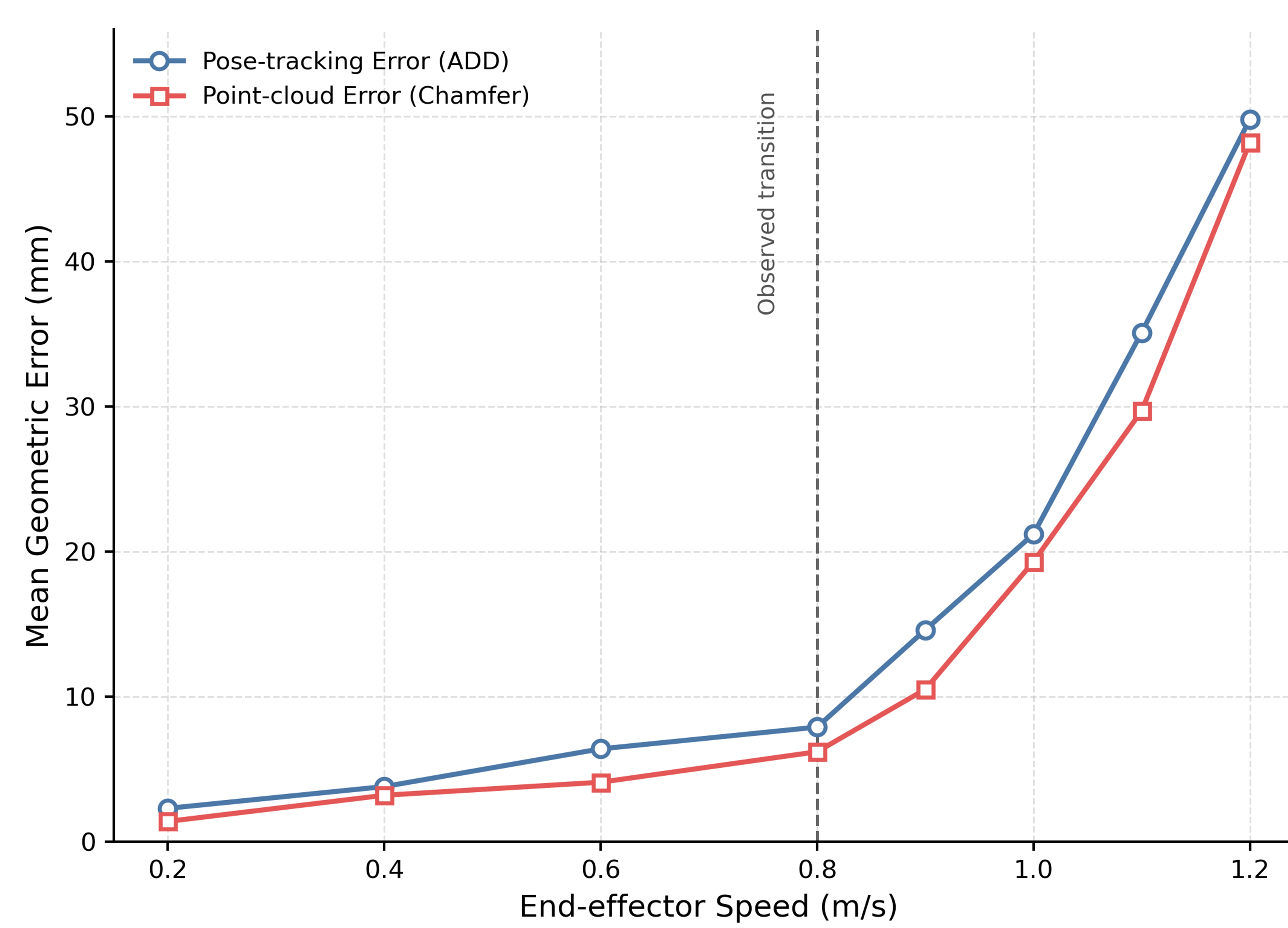}
	\caption{Pose and point-cloud errors at different speeds.}
	\label{fig18}
	\vspace{-10pt}
\end{wrapfigure}

\subsection{Pose-Tracking Analysis under Motion Speed}
\label{Pose-Tracking Analysis under Motion Speed}
The preceding Dream-Mode robustness analysis (Sec.~\ref{Dream-Mode Robustness Analysis}) shows that short-term hand--object occlusion can be partially handled through latent prediction. Beyond occlusion, object motion speed is another important factor affecting the quality of interaction-point-cloud estimation. To evaluate this effect, we replay training trajectories at different end-effector speeds and measure the object pose-tracking error (ADD) and point-cloud error (Chamfer distance), both in millimeters. As shown in Fig.~18, both errors remain low up to $0.8\,\mathrm{m/s}$, where the pose and point-cloud errors are approximately $8$ mm and $6$ mm, respectively. Beyond $0.8\,\mathrm{m/s}$, both errors increase rapidly and approach $50$ mm at $1.2\,\mathrm{m/s}$, indicating degraded rotation estimation and point-cloud alignment under fast motion. Since proximity estimation relies on accurate hand--object relative geometry, such tracking errors directly reduce the reliability of proximity cues. Therefore, the current system is better suited to moderate-speed contact-rich manipulation, while highly dynamic motions remain limited by real-time pose-estimation accuracy.

\FloatBarrier

\section{Real-World Experiments}
\label{app:real_world}

\subsection{Experimental Setup and Task Description}
\label{app:real_setup}

\begin{figure}[!t]
	\centering
	\captionsetup{skip=4pt}
	\includegraphics[width=1.0\textwidth]{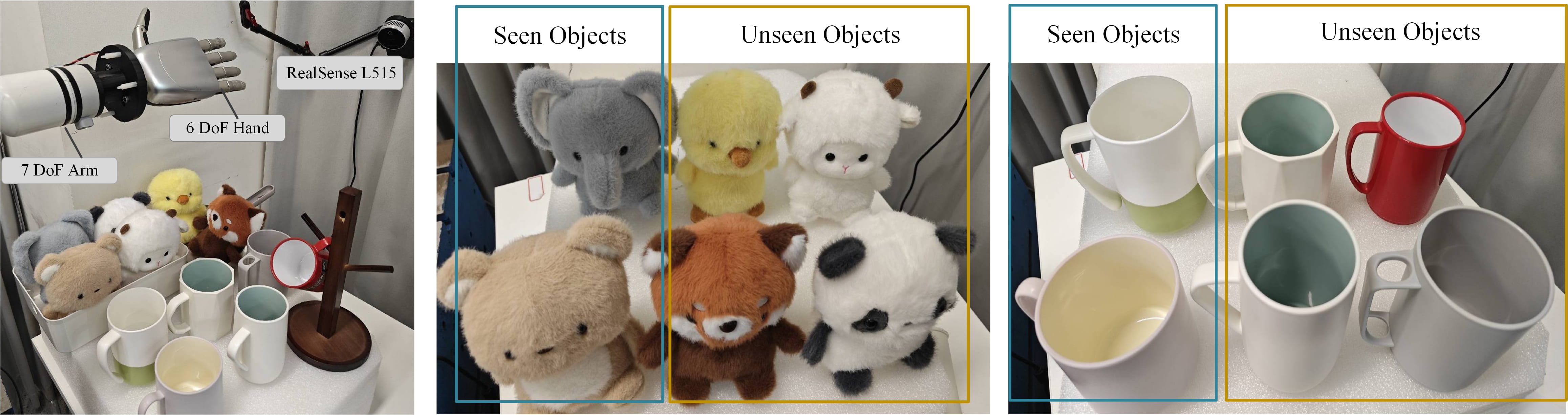}
	\caption{Real-world platform and evaluation objects. 
		We use a RealMan RM75B arm with a 6-active-DoF dexterous hand and a fixed Intel RealSense L515 camera, and evaluate on seen and unseen objects with diverse shapes and sizes.}
	\label{fig:real_platform}
	\vspace{-10pt}
\end{figure}

Real-robot experiments evaluate ProxiDex under real perception noise, cross-object variation, and manipulation disturbances. The platform consists of a Realman RM75B 7-DoF arm, an Intel RealSense L515 RGB-D camera, and two dexterous hands, as summarized in Fig.~\ref{fig:real_platform}. A 6-DoF CasBot-P1L hand is used for standard tabletop tasks, while a 20-DoF Wuji Hand V1 is used for more contact-rich manipulation. The policy must control not only the global arm pose but also finger closure timing and local contact state. The external RGB-D camera provides object pose tracking, and the reconstructed object point cloud is combined with robot forward kinematics to generate the hand-object interaction point cloud.

The three real-world tabletop tasks are \textbf{Pick}, \textbf{Pinch}, and \textbf{Sweep}. Pick requires the robot to grasp a tabletop object with multiple fingers and place it into a basket, testing grasp stability, transport, and release timing. Pinch requires the robot to form a small-area thumb-index contact and transfer the target object to the desired location, emphasizing fingertip alignment and stable closure. Sweep requires the robot to push a target object into a dustpan, grasp the dustpan handle, and pour the object into a container, involving pushing, tool grasping, and secondary contact transfer. In addition, Wuji Hand V1 is used for two contact-rich tasks involving sustained finger--object interaction and object reorientation. Together, these tasks cover whole-hand grasping, precise fingertip contact, and sustained in-hand interaction.

Demonstrations are collected using the VR teleoperation system described in Appendix~\ref{app:vr_teleoperation}, with the operator controlling the arm and hand through Quest 3S. For each task, we collect $50$ high-quality demonstrations while varying object initial positions, orientations, and some object types to improve adaptation to different geometric configurations. During point-cloud visualization, object pose estimates may exhibit mild jitter. This perturbation does not break action learning; instead, it increases observation variation in the training data and makes both the policy and the dynamics checker more robust to mild perception errors in deployment.

Fig.~\ref{fig:real_execution_point_clouds} visualizes real-world executions and their reconstructed interaction point clouds. The proximity responses remain concentrated near active hand--object contact regions, such as the grasping fingers in Pick, the thumb-index contact in Pinch, and the tool-object contact transfer in Sweep. This qualitative behavior links the real-world setup back to the geometric proximity representation in Appendix~\ref{app:geometric_proximity} and motivates the evaluation protocol below.

\subsection{Evaluation Protocol}
\label{app:real_protocol}

We use three test settings to evaluate generalization and robustness. \textbf{In Distribution} uses the same objects and similar scene configurations as data collection, measuring task completion under standard conditions. \textbf{Unseen Objects} uses objects that do not appear in training and differ in shape, size, or surface appearance, testing generalization to object geometry changes. \textbf{Perturbation} introduces complex tabletop backgrounds, clutter, short-term visual occlusion, or mild pose perturbations during execution, evaluating robustness under unstable perception.

Success criteria consider not only the final object location, but also contact stability and pose consistency during execution. For Pick, the robot must stably grasp the target and place it into the basket. For Pinch, it must form a stable thumb-index pinch and transfer the target to the specified area.  For Sweep, it must sweep the target into the dustpan and complete pouring into the target container by grasping the dustpan. We evaluate ProxiDex under three test settings: standard execution, cross-object generalization, and disturbance robustness.

\subsection{Failure Case Analysis}
\label{app:failure_cases}
\begin{wrapfigure}{r}{0.4\textwidth}
	\vspace{-8pt}
	\centering
	\includegraphics[width=0.38\textwidth]{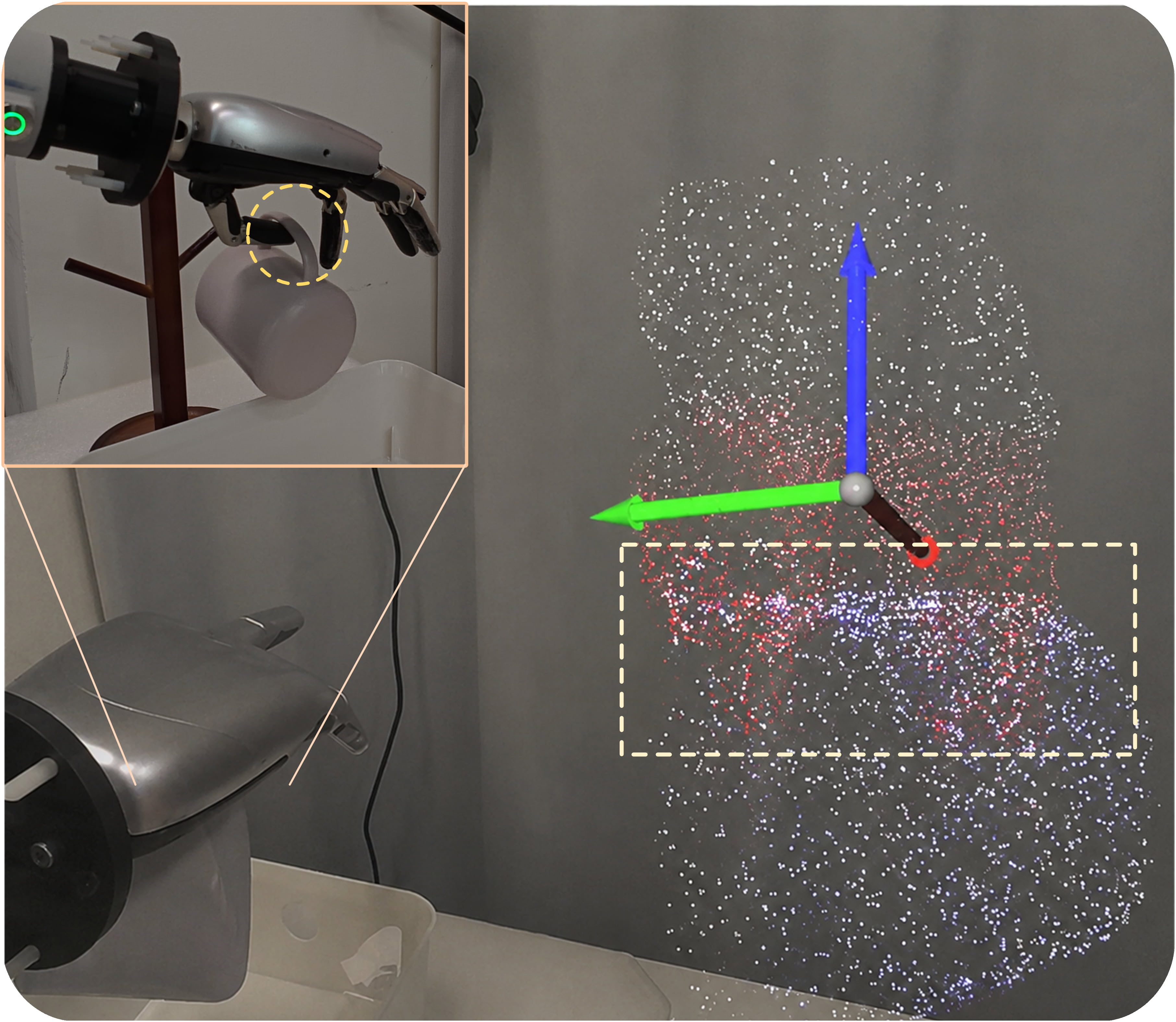}
	\caption{Failure cases under fine interaction. Finger alignment may occasionally be inaccurate, and object pose estimation may show mild jitter during contact-rich manipulation.}
	\label{fig14}
	\vspace{-10pt}
\end{wrapfigure}
The results above collectively demonstrate the accuracy and robustness of ProxiDex, while Fig.~\ref{fig14} illustrates the remaining limitations of ProxiDex in fine-grained manipulation. Although ProxiDex achieves a practical balance between low-cost data collection and real-world deployment, its current proximity representation is still mainly geometry-driven and cannot fully capture precise fingertip force distribution. As shown in the zoomed failure case, the middle finger and thumb may occasionally fail to align symmetrically with the object during closure, resulting in unbalanced contact forces. In addition, during the cup-pinching task, the cup may exhibit slight shaking during detachment and transfer. This motion, together with partial hand-object occlusion, can reduce the accuracy and temporal stability of object pose estimation, thereby perturbing the interaction point cloud and affecting the stability of proximity-based feedback.

Future work can incorporate data gloves or lightweight tactile sensors to obtain higher-precision fingertip contact information, and use contrastive learning to align it with the geometric proximity representation proposed in this work. This would preserve the low-cost and cross-platform advantages of ProxiDex while improving force awareness and finger coordination in fine manipulation.

\begin{figure}[!htbp]
	\centering
	\captionsetup{skip=4pt}
	\includegraphics[width=1.0\textwidth]{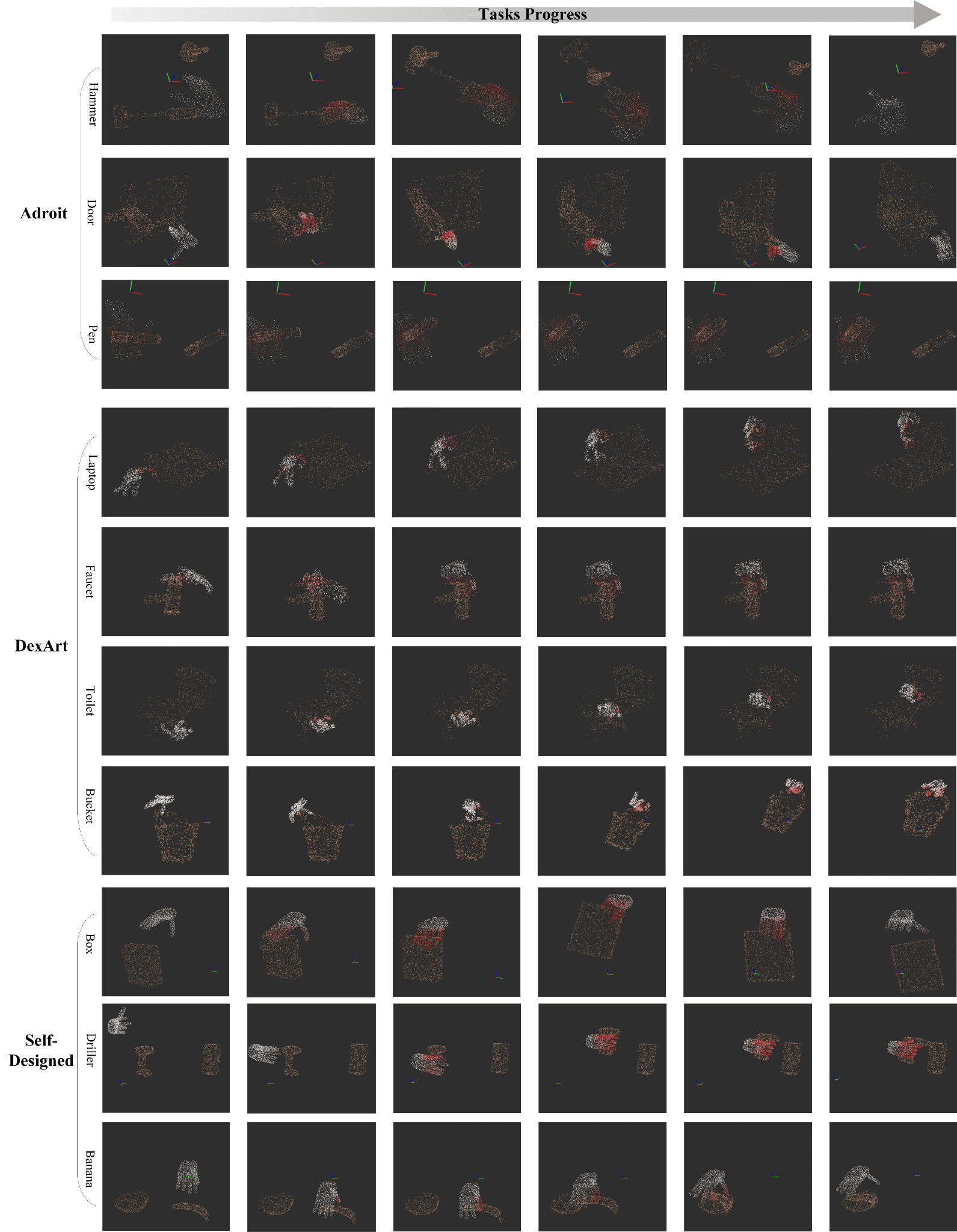}
	\caption{Visualization of interaction point clouds in simulation. 
		Across Adroit, DexArt, and self-designed tasks, the processed hand--object point clouds preserve fine-grained interaction geometry, with red regions indicating stronger proximity responses along the task progress.}
	\label{fig:simulation_point_clouds}
	\vspace{-10pt}
\end{figure}

\begin{figure}[!t]
	\centering
	\captionsetup{skip=4pt}
	\includegraphics[width=1.0\textwidth]{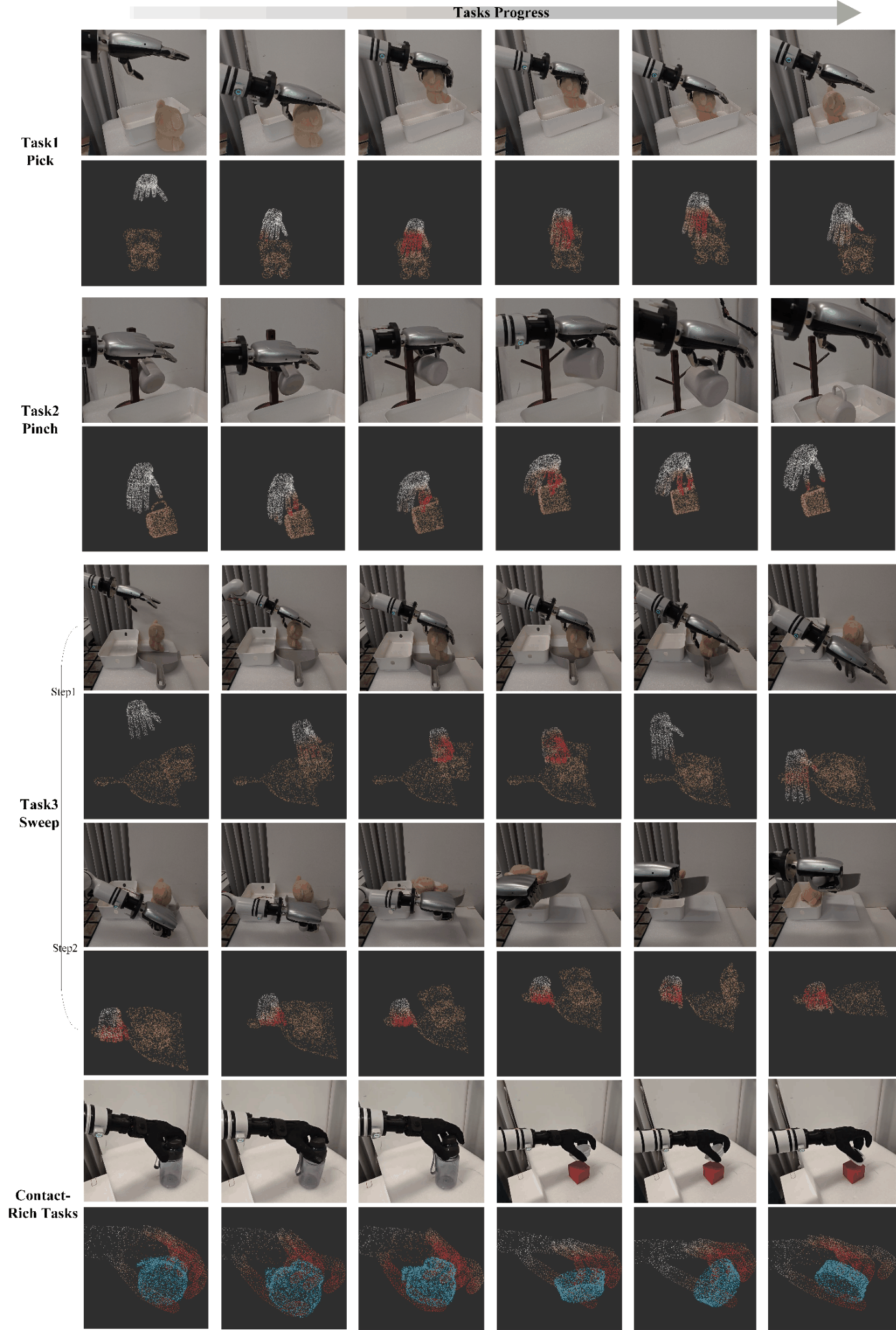}
	\caption{Visualization of real-world executions and the corresponding interaction point clouds. 
		The reconstructed point-cloud sequences closely match the camera observations, while the color-coded proximity responses highlight local hand--object interaction regions during manipulation.}
	\label{fig:real_execution_point_clouds}
	\vspace{-10pt}
\end{figure}

\end{document}